\documentclass[runningheads]{llncs}

\usepackage{eccv}

\usepackage{eccvabbrv}

\usepackage{graphicx}
\usepackage{booktabs}

\usepackage[accsupp]{axessibility}  

\usepackage{hyperref}
\usepackage{multirow}

\usepackage{orcidlink}
\usepackage{algorithm,algcompatible}
\usepackage{algpseudocode}

\begin{document}

\title{Towards Robust Event-based Networks for Nighttime via Unpaired Day-to-Night Event Translation} 

\titlerunning{Unpaired Day-to-Night Event Translation}


\newcommand\CoAuthorMark{\footnotemark[\arabic{footnote}]}
\author{Yuhwan Jeong\orcidlink{0009-0002-0279-146X}\thanks{Equal contribution.} \and
Hoonhee Cho\orcidlink{0000-0003-0896-6793}\protect\CoAuthorMark \and
Kuk-Jin Yoon\orcidlink{0000-0002-1634-2756}}

\authorrunning{Jeong et al.}

\institute{Korea Advanced Institute of Science and Technology\\
\email{\{jeongyh98, gnsgnsgml, kjyoon\}@kaist.ac.kr}\\
}

\maketitle

\begin{abstract}
Event cameras with high dynamic range ensure scene capture even in low-light conditions.
However, night events exhibit patterns different from those captured during the day.
This difference causes performance degradation when applying night events to a model trained solely on day events.
This limitation persists due to a lack of annotated night events. To overcome the limitation, we aim to alleviate data imbalance by translating annotated day data into night events.
However, generating events from different modalities challenges reproducing their unique properties.
Accordingly, we propose an unpaired event-to-event day-to-night translation model that effectively learns to map from one domain to another using Diffusion GAN.
The proposed translation model analyzes events in spatio-temporal dimension with wavelet decomposition and disentangled convolution layers. We also propose a new temporal contrastive learning with a novel shuffling and sampling strategy to regularize temporal continuity.
To validate the efficacy of the proposed methodology, we redesign metrics for evaluating events translated in an unpaired setting, aligning them with the event modality for the first time.
Our framework shows the successful day-to-night event translation while preserving the characteristics of events.
In addition, through our translation method, we facilitate event-based modes to learn about night events by translating annotated day events into night events. Our approach effectively mitigates the performance degradation of applying real night events to downstream tasks. The code is available at \url{https://github.com/jeongyh98/UDNET}.

\end{abstract}

\section{Introduction}
Deep learning models trained on data from any sensors are biased and optimized for the training dataset. Consequently, they often face a drop in performance when confronted with a new domain. 
Acquiring new labels for every new domain is challenging. Therefore, research on methods to transfer style with translation~\cite{gatys2016image,anokhin2020high,zheng2020forkgan,schutera2020night,fan2023learning, 10323514} is actively being pursued.

\begin{figure}[t]
    \centering
    \includegraphics[width=0.99\linewidth]{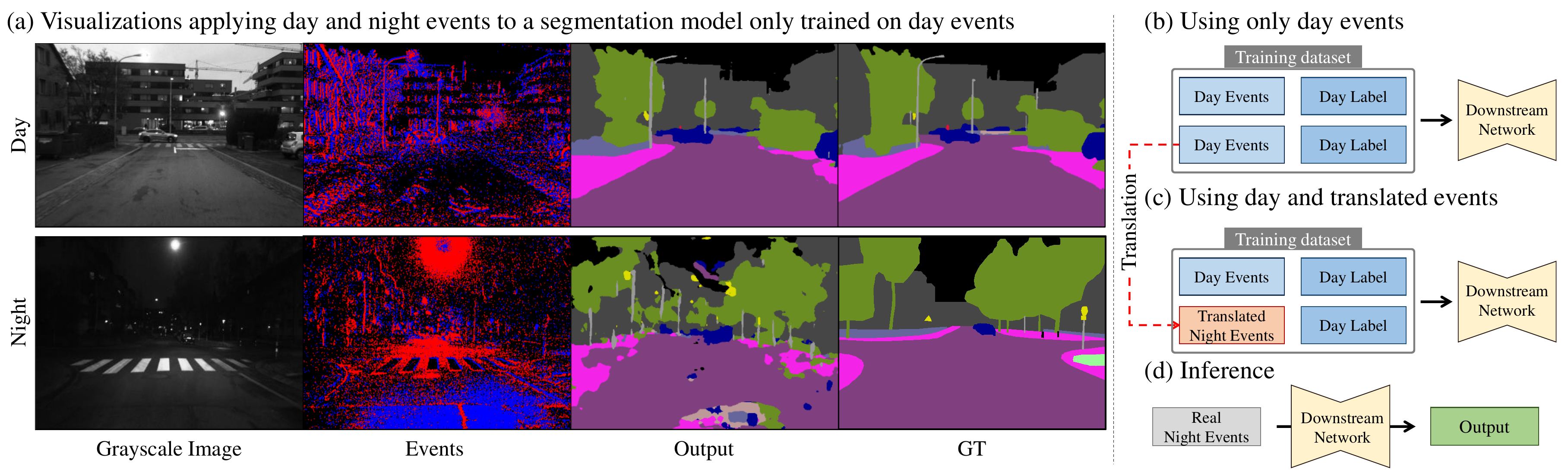}
    \caption{(a) The results of a segmentation model~\cite{sun2022ess} trained on day event data. Day events inference yields comparable results to the ground truth, while night events inference is unsatisfactory. (b) Due to a scarcity of nighttime data, the network is trained solely on day events and their labels. (c) We translate day events into night events to train the network and then (d) improve performance while applying real night events.}
    \label{fig:intro_night_model}
\end{figure}

This performance drop also occurs in event-based networks~\cite{sun2022ess, cho2023learning, cho2023non}.
An event camera~\cite{4444573, 9063149, gallego2020event} records light intensity levels between the current and subsequent frame for each pixel.
Event cameras can sense the night more effectively than conventional cameras~\cite{shi2023even, cadena2023sparse, zhang2020learning, zhu2018ev, cho2022selection, liu2023low, cho2024tta}. Due to their characteristics, event cameras are highly suitable sensors for autonomous driving in both day and night environments. An event-based model~\cite{gehrig2023recurrent, zhang2023frame, alonso2019ev, kim2024frequency, cannici2019asynchronous, tulyakov2019learning, wang2019ev, cho2023label, wu2022video, cho2022event} trained on day event data can yield high-quality results for day events. However, they have limitations when dealing with unseen night events (see Fig.~\ref{fig:intro_night_model}). Unlike day events, night events include excessive activation of one polarity due to artificial lighting, inherent noise, and edge blurring. Therefore, we focus on autonomous driving street scenes with abundant artificial lighting and numerous structures, where excessive activation and edge blurring can become severe.

The most effective approach to address this issue is to use all data from each domain during the training.
However, obtaining night event annotations is challenging compared to the abundance of available daytime data.
To mitigate these challenges and produce robust outputs regardless of day or night, we focus on day-to-night translation.
The unique attribute of data translation lies in its lack of pairing~\cite{huang2021unsupervised} because a strict alignment of the sensor output between day and night is challenging.
As what is lacking for our goal is labels for night events, our task outputs night events with an unpaired manner.
When striving for unpaired translation with the night events-like output, an input for carrying out this task should involve utilizing data captured during the daytime.

One of the considerations is conventional sequential images (\ie,~video).
Consecutive frames can produce events through the intensity differences between adjacent time points. Nonetheless, events encompass multiple attributes, such as spatio-temporal features and temporal continuity in pixels, that differ from an image modality.
These differences limit the quality of generated events from images like EventGAN~\cite{zhu2021eventgan} and CMDA~\cite{xia2023cmda}.
Image-based methods encounter difficulties constructing temporal flow for instantaneous events and struggle to produce fast-moving objects properly.
As a result, bypassing the creation from images and commencing the translation from events is the path to successful problem-solving. Therefore, we initiate the translation from events.

To this end, we need to extract features effectively from day events and translate them to night events.
Typically, the translation process utilizes various generation models.
GAN~\cite{goodfellow2014generative,zhu2017unpaired,park2020contrastive} allows us an unpaired translation; however, it may focus only on specific modes, failing to diversify scenes of events.
Straightforward diffusion model~\cite{ho2020denoising, song2020denoising} is not inherently suitable for the unpaired.
Lastly, Diffusion GAN~\cite{xiao2022tackling, kim2023unpaired}, utilizing a discriminator with a diffusion process, enhances the overall stability and diversity during an unpaired translation.

Therefore, we propose our unpaired event-to-event day-to-night translation, which successfully reproduces the style of events based on Diffusion GAN~\cite{kim2023unpaired}.
However, directly applying models proposed for the image modality to the event modality disrupts the consistency along the temporal axis of the event representation~\cite{liu2018adaptive, duan2021eventzoom, weng2022boosting}. 
In other words, while methods for an image modality can adequately handle changes in style at the feature level, they do not address the temporal continuity of events.
To solve these issues, we propose a novel and practical approach, temporally shuffling contrastive regularization for a Diffusion GAN model designed for events.
To create negative samples, we randomly shuffle the order of the temporal dimension in the event representation. Conversely, the original representation is a positive sample, following contrastive learning setup~\cite{Chen2020ASF}.
Since typical convolution operations aggregate all channels into a single convolution without temporal separation, we propose a temporally disentangled encoder to prevent the saturation in the temporal dimension and enhance the contrastive learning method.
Furthermore, recognizing that events primarily activate in the high-frequency domain, especially the style of night events existing in the high-frequency domain, we introduce a new method to process high-frequency information through wavelet decomposition.

Through our coherent temporally shuffling contrastive regularization and temporally disentangled encoder, we successfully maintain the temporal consistency of events and effectively handle style changes for night scenes using wavelet decomposition. Particularly, methods originally intended for the image modality show instability in the event modality, overly focusing on style changes. However, our approach adequately reflects the properties of events that change over time.
To quantify produced events, we redesign image metrics into new metrics.
To demonstrate the strength of our method, we compare it with other methods in several datasets. Furthermore, we conduct several downstream tasks to show the practicality of this approach. Training models on datasets generated using our method resulted in the most significant performance enhancement in these downstream tasks, affirming the effectiveness of our approach.

\section{Related Works}

\subsection{Nighttime Event Data Generation}
Due to distinct distributions of day and night events, applying night events to networks trained on day event data leads to a performance drop.
In this context, the term ``distribution'' refers to statistical characteristics of values within the events 
To reduce the drop, generation methods have been applied.

\noindent
\textbf{Frame-based event generation} attempts to adapt to the night domain or generate night events. Notably, EventGAN~\cite{zhu2021eventgan} proposes a method that generates events from two sequential images. Its quality follows the quality of the input images. Therefore, a scarcity of information, such as night driving scenes, leads to difficulties in representing essential structures and key details.
To address these issues, CMDA~\cite{xia2023cmda} suggests generating night events from sequential day images leveraging the computation of intensity differences among discrete images. Nevertheless, this leverage deviates from the actual event distribution, missing considerations of the temporal axis in event representations~\cite{zhu2019unsupervised,wang2019event,maqueda2018event}.
The approaches from images do not adequately reflect the temporal characteristics of events and struggle with generating realistic events.
Most of these approaches have outputs in predefined representations of events, such as voxels~\cite{zhu2019unsupervised} or histogram~\cite{duan2021eventzoom, weng2022boosting} forms, and we also follow them, with both the input and output of events being in histogram form.

\noindent
\textbf{Event simulators}~\cite{rebecq2018esim, gehrig2020video, hu2021v2e, lin2022dvs, gu2023reliable} are another way to creating synthetic events. One of the biggest advantages of the simulation is generating a raw event stream in the same form as the actual one. These researches demonstrate robust performance in areas that have been studied. However, it exhibits relatively weak performance at night and with changing domains like new cameras. The current execution of simulation research exhibits some misalignment with generating night events with day data in various situations. Therefore, we propose the unpaired translation starting from day events to create the target, night events.

\subsection{Generation Model for Unpaired Translation}
\noindent
\textbf{Generative Adversarial Networks (GANs)}
involve training a generator to learn the data distribution of the training set. While simultaneously training a discriminator to distinguish between generated outputs and real data.
CycleGAN~\cite{zhu2017unpaired, liu2017unsupervised} uses cycle-consistency to construct a two-sided mapping function between two domains. However, to maintain cycle-consistency, two pairs of generator-discriminator are required.
To simplify the network, one-way translation algorithms~\cite{amodio2019travelgan, benaim2017one, fu2019geometry, park2020contrastive, song2023shunit} have been considered.
Despite impressive improvements, several problems, such as mode collapse, remain for GANs.

\noindent
\textbf{Diffusion Models}~\cite{ho2020denoising}, inspired by the movement of molecules, generate images through a stochastic process. It consists of forward diffusion and reverse denoising processes following Markov chains to restore the original inputs from the Gaussian noise.
Based on this, additional approaches emerge, including the utilization of score functions~\cite{song2021scorebased}, the development of fast sampling techniques~\cite{song2021denoising}, the relocation of processes to latent space~\cite{rombach2022high}.
Meanwhile, UNIT-DDPM~\cite{sasaki2021unit}, a two-sided network, offers the potential for success in unpaired translation; however, it requires substantial computational resources.

\noindent
\textbf{Diffusion GAN}~\cite{xiao2022tackling} is proposed to combine the advantages of GAN and diffusion models to overcome their weaknesses.
In contrast to general diffusion research conducted in the paired setting, there is ongoing research in an unpaired manner. Especially, UNSB~\cite{kim2023unpaired} has proposed a new approach to unpaired image-to-image translation by combining a discriminator and a Schrödinger Bridge diffusion model~\cite{wang2021deep}.
While it is possible to import UNSB to translate events, the characteristics of output events are not preserved. To make the output realistic, we propose a new method to facilitate modeling spatio-temporal properties and natural noise of night events. 

\section{Method}

\subsection{Preliminary}

\textbf{Event Representation.}
\label{sec:representation}
The event camera records specific changes in times and locations, resulting in sparsely distributed streams as $\mathcal{E} = \mathrm{\Sigma}e_k(t)$.
A single $k$-th event of the stream is $e_k(t) = (x_k, y_k, t_k, p_k)$ where $x_k, y_k$ are event-triggered locations, $t_k$ is a trigger time, and $p_k$ is its polarity, $\pm1$.
To make events as a 3D volume, we stack events with each polarity from a single sequence of the event stream, $\mathcal{E}(t_k, t_k + \Delta t)$. We set polarity and temporal bins as channels to create a series of event histograms~\cite{liu2018adaptive, duan2021eventzoom, weng2022boosting}, $E \in \mathbb{R}^{(2\times B) \times H \times W}$.
Previous work~\cite{xia2023cmda} did not consider the temporal dependency of events and disregarded polarity, limiting the channel size to 1. 
In contrast, we design experiments with bin sizes of $B$ of 1, 3, and 8 to avoid the loss of temporal information for diverse applications.
We denote our representations of day and night events as $E_D$ and $E_N$.

\noindent
\textbf{Connecting two arbitrary distributions with Schrödinger bridge.}
In our problem, we have two distributions, which correspond to $E_D$ and $E_N$, denoted as \(\pi_0\) and \(\pi_1\)  on $\mathbb{R}^d$. We aim to find the optimal transport path from \(\pi_0\) to \(\pi_1\) using the Schrödinger Bridge problem (SBP)~\cite{de2021diffusion}.
The SBP explores the random process $\{X_t :t \in [0, 1]\}$ which follows: 
\begin{equation}
    \mathbb{Q}^{*} = \underset{\mathbb{Q} \in P(\Omega)}{\mathrm{argmin}} \mathbb{D}_{KL}(\mathbb{Q}\parallel \mathbb{W}_{\tau}),~\text{s.t.}~\mathbb{Q}_0 = \pi_0, \mathbb{Q}_1=\pi_1
    \label{eqn:Q_*}
\end{equation}
where $\mathbb{W}_{\tau}$ is the Wiener measure with variance $\tau$, $\Omega$ is the space of continuous functions on time interval $t$, and $P(\Omega)$ denotes the space of probability measure on $\Omega$. 
To discover $X_t$ that satisfies the Schrödinger Bridge, $\left\{ X_{t}\right\}$ $\sim$ $\mathbb{Q}^{*}$, we use the stochastic control formulation~\cite{dai1991stochastic} in the form of an SDE.
According to Schrödinger Bridge continuous flow matching (CFM)~\cite{tong2023improving}, while the initial condition $X_{0}$ is given, the conditional probability can plan the marginal process $X_{t}$.
Considering the entropy-regularized optimal transport problem~\cite{hernandez2012discrete} with two distributions $\pi_0$ and $\pi_1$, SBP should be solved by:
\begin{equation}
    \mathbb{Q}_{01}^{*}=\underset{{\pi \in\Pi(\pi_0,\pi_1)}}{\operatorname{argmin}}\mathbb{E}_{(X_0,X_1) \sim \pi}\left [ {\left\| X_0 - X_1 \right\|}^2 \right ]-2\tau H(\pi),
    \label{eqn:final}
\end{equation}
where $H$ is the entropy function of transport distribution $\pi$.
We formulate the path of two distributions recursively by the Markov chain decomposition: $p(\{X_{t_{m}} \}) = p(X_{t_{M}}|X_{t_{M-1}})\cdots \, p(X_{t_{1}}|X_{t_{0}}) p(X_{t_{0}}).$
To seek the SB process $p$, we define $q_{\phi_i}$ as a conditional distribtuion parametrized by a DNN with parameter $\phi_i$. Then, to stabilize our path optimize $q_{\phi_i}$, we utilize the below losses while applying restricted interval $[t_i, 1]$ as following~\cite{kim2023unpaired}:

\begin{equation}
    \begin{gathered}
        \underset{\phi_i}{\textnormal{min}}\;\mathfrak{L}_{*}(\phi_i, t_i) = \mathbb{E}_{q_{\phi_{i}}(X_{t_i}, X_1)} \left [ \left\| X_{t_i}-X_1\right\|^2\right ] -2\tau(1-t_i)H(\phi_i (X_{t_i}, X_1)) \\
        \textnormal{s.t.}\;\;\mathfrak{L}_{\textnormal{Adv}}(\phi_i, t_i)=\;D_{\textnormal{KL}}(q_{\phi_i}(X_1)\parallel p(X_1)) = 0
        \label{eqn:final_const}
    \end{gathered}
\end{equation}

\begin{figure}[!t]
    \centering
    \includegraphics[width=0.99\linewidth]{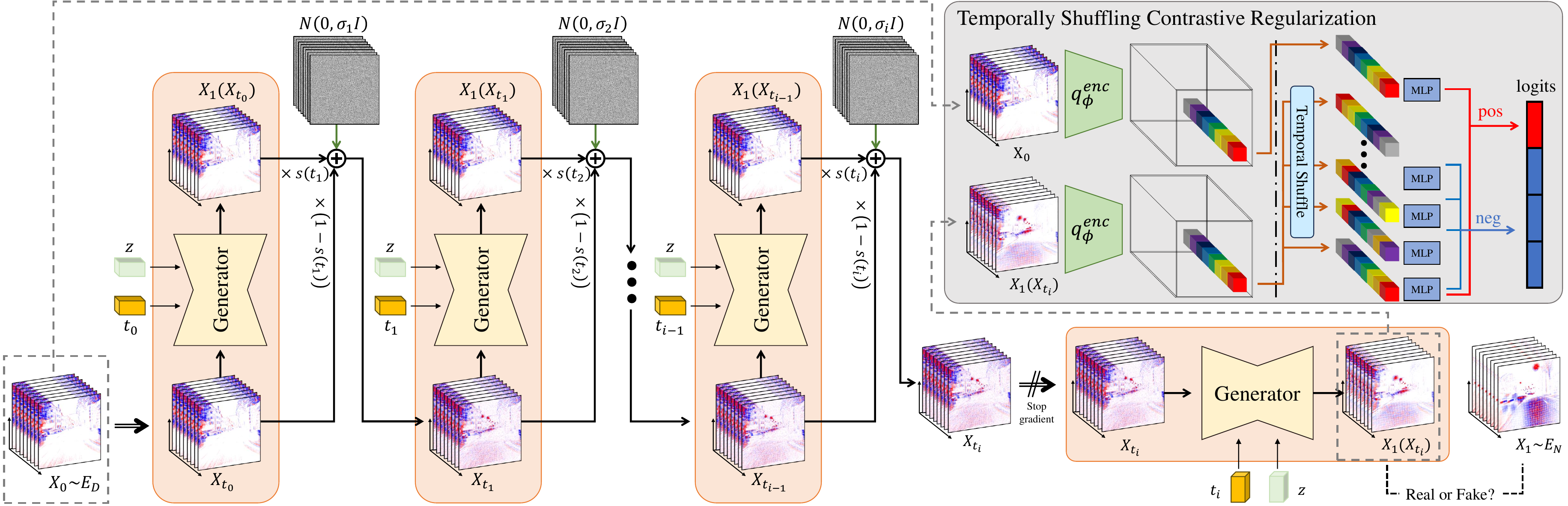}
    \caption{Overall process of our proposed method. To build an unpaired day-to-night event translation model, we train a generator using various constraints. To preserve the spatio-temporal property of events, we propose a novel temporally shuffling contrastive regularization with spatial contrastive regularization.}
    \label{fig:framework1}
\end{figure}

\subsection{Overall Framework}
Our goal is to build a framework to translate day events to night events in an unpaired manner, $\hat{E}_{N} = G_{D\rightarrow N}(E_D)$ where $G_{D\rightarrow N}$ is a day to night event translator, \ie,~a generator.
Our proposed framework is briefly visualized in Fig.~\ref{fig:framework1}. We incorporate training strategies of UNSB~\cite{kim2023unpaired}, a kind of Diffusion GAN, into our framework.
The whole process is divided into two steps: a sampling step and a training step.
In a sampling step, we construct the Schrödinger process, $\{X_t\} \sim \mathbb{Q}^{*}$, that goes from a day event distribution to a night event distribution. The discrete path is connected with several intermediate timestamps $\{t_i\}$ where $i = 0, 1, \cdots, M-1$ with $0=t_0 \leq t_i < t_{i+1} \leq t_M = 1$. To follow the path, we first sample $X_{t_0}$ from $E_D$ and $X_{1}$ from $E_N$. 
Passing the sample through the generator produces an intermediate sample that slightly moves towards $E_N$.
We use the UNet-like~\cite{ronneberger2015u} generator to translate events.
Then, we interpolate the input and output of the generator with the following equation: $p(X_{t} | X_{t_{0}}, X_{1}) = \mathcal{N}(X_t|s(t)X_{1}+(1-s(t))X_{t_0}, s(t)(1-s(t))\tau(1-t_0)\,I)$,
where $s(t)=(t-t_0)/(1-t_0)$, to obtain a next intermediate sample $X_{t_1}$. We repeat the noising step iteratively to generate a final sample $X_{t_i}$.
For the training step, we train the generator with a pair of the final sample $X_{t_i}$ and $X_1$ while the sampling step makes the sample converge to the target distribution.
The path between the two samples follows the Schrödinger Bridge and is stabilized by constraints based on Schrödinger conditions. 
With Eq.~(\ref{eqn:final_const}), we formulate the loss function willing to follow the optimal path as follows: $\underset{\phi_i }{\textnormal{min}}\;\mathfrak{L}(\phi_i, t_i) = \mathfrak{L}_{Adv}(\phi_i, t_i) + \lambda_{SB, t_i}\mathfrak{L}_{SB}(\phi_i, t_i).$
As $ t \rightarrow 1$ and training goes on, the generator converts the noisy sample to the night events.
Using only SB entropy and optimal path loss, the sampled day event $E_D$ is translated into night events $\hat{E}_N$; however, translated night events may not preserve structural information of the input because sampled two inputs, $E_D$ and $E_N$, are unpaired.
To account for the spatio-temporal characteristics of events, we add a spatial contrastive constraint to preserve structural information and a novel temporally shuffling contrastive constraint to maintain the continuity:
\begin{align}
    \mathfrak{L}(\phi_i, t_i) = \mathfrak{L}_{Adv}(\phi_i, t_i) + \lambda_{SB, t_i}\mathfrak{L}_{SB}(\phi_i, t_i) + \nonumber\\ \lambda_{SC, t_i}\mathfrak{L}_{SC}(\phi_i, t_i) + \lambda_{TC, t_i}\mathfrak{L}_{TC}(\phi_i, t_i).
    \label{eqn:total_loss}
\end{align}
\noindent
$\mathfrak{L}_{Adv}$ is computed between $X_1(X_{t_i})$ and $X_1$, for $\mathfrak{L}_{SB}$, $X_{t_i}$ and $X_1(X_{t_i})$, and for $\mathfrak{L}_{SC}, \mathfrak{L}_{TC}$, $X_0$ and $X_1(X_{t_i})$. Since we use a time-conditional network, for practicality, we apply the same parameter, $\phi$, to all instances of $\phi_i$.

\begin{figure}[t]
    \centering
    \includegraphics[width=1\linewidth]{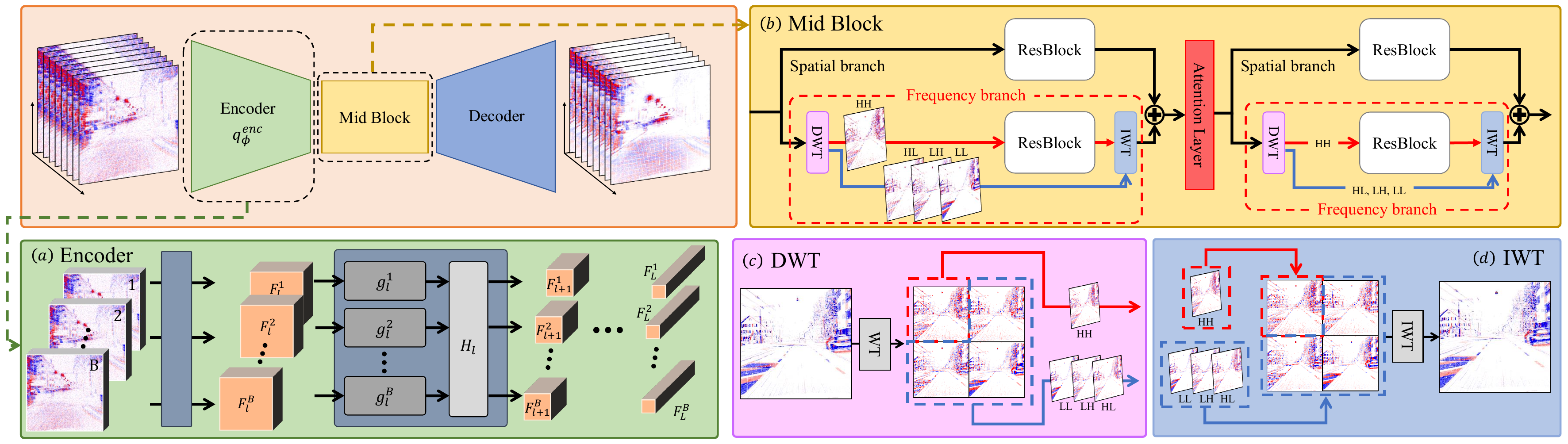}
    \caption{Details about our proposed generator. In the encoding process, we disentangle channels and merge them by their original time bin. Channels from each bin pass through the disentangled block, $\mathcal{G}_l$, and merge by the global block, $\mathcal{H}_l$, as shown in (a). }
    \label{fig:framework2}
\end{figure}

\noindent
\textbf{Temporally Disentangled Encoder.}
\label{Sec:TDE}
Following the footsteps of previous diffusion models~\cite{kim2023unpaired, ho2020denoising}, we design a network using convolution layers. However, standard convolution layers do not explicitly encode temporal information; instead, they merge information from different time intervals into a single channel through convolution operations.
Unlike images, where information from each channel is from the same moment, events store temporal information within the channel dimensions of the input tensor.
As a result, each of the bins contains information from distinct time intervals, and this difference is also reflected spatially in the pixels. To effectively encode events by preserving temporal order for each spatial location, we propose a temporally disentangled encoder, $q_\phi^{enc}$. 

As shown in Fig.~\ref{fig:framework2}, the temporally disentangled encoder, $q_\phi^{enc}$, consists of two components at each scale level, $l \in \{1, 2, \dots, L\}$. 
One is a disentangled block responsible for encoding each temporal information separately without sharing, and the other is a global block that allows interaction with the overall information along with the temporal dimension.  
More specifically, we first divide the channels of features $F_l \in \mathbb{R}^{H_l\times W_l \times C_l}$ into time bins $B$. Each feature group has $C_l/B$ channels. With the temporally disentangled block $\mathcal{G}_l^b$ for the $b$-th bin at the scale level $l$, we encode the features as: $\tilde{F}_l^b = \mathcal{G}_l^b(F_l^b), \; b \in {1,2,\dots, B}$ where $F_l^b, \tilde{F}_l^b \in \mathbb{R}^{H_l\times W_l \times C_l/B}$.
During this process, the network remains completely isolated from different temporal information along the temporal axis.
This enables the network to learn about differences over time implicitly. Instead, we only allow interaction along the channel dimensions when downsampling operations through the global block $\mathcal{H}_l$, as follows: $F_{l+1} = \mathcal{H}_l([\tilde{F}^1_l, \tilde{F}^2_l, \dots, \tilde{F}^B_l]).$

\noindent
\textbf{Wavelet-based Decomposed Bottleneck Block.}
The networks exhibit a bias toward prioritizing the tuning of low-frequency features first~\cite{rahaman2019spectral}. High-frequency features are less robust and significantly influenced by changes in low-frequency elements.
Various generative models~\cite{gal2021swagan, wang2022fregan} have attempted to decompose features in the frequency domain. Through decomposition, they focus on fitting low-frequency components that affect the style the most.
Based on these analyses, we introduce wavelet decomposition to transfer the style of events effectively. We find that the overall structure is contained within the low frequency, while noises corresponding to the style of the night are predominantly distributed in the high frequency.
Therefore, we are motivated by the idea that the style of event data can be better transferred through processing in the high-frequency domain to reproduce the nature of event noises (More about the motivation is described in the supplementary materials).

In general, diffusion models encode features first and generate high-level features through attention~\cite{rombach2022high} at the lowest resolution.
We also follow the same scheme, but we propose a method of concentrating features in both spatial and frequency domains to model the noise of night events competently.
As shown in (b) of Fig.~\ref{fig:framework2}, we design the mid block by splitting it into two branches: spatial and frequency.
The spatial branch is a usual residual block. On the other hand, in the frequency branch, we decompose input features $F$ into low $F_{LL, HL, LH}$ and high-frequency components $F_{HH}$ through Wavelet Transform (WT)~\cite{zhang2019wavelet}.
We maintain the low-frequency components through residual connections to preserve the overall structure.
For the high-frequency domain, we manipulate it with the resblock to enhance the awareness of night event noise.
The original low-frequency feature map and the processed high-frequency subband are reverted to their original space through an Inverse Wavelet Transform (IWT).
Following that, the outcome from the frequency branch is merged with the output of the spatial branch.

\noindent
\textbf{Temporally Shuffling Contrastive Regularization.}
Events with a smaller bin index should precede events with a larger index in their temporal sequence. One of our goals is that translated night events maintain the temporal order of day events.
To enhance the temporal order and continuity, we apply an extra regularization method. We propose a new temporal regularization based on a contrastive learning method~\cite{le2020contrastive} that aims to associate positive information related to a given query closer while pushing negatives farther away as:
\begin{align}
 \mathfrak{L}_C\left(\boldsymbol{v}, \boldsymbol{v}^{+}, \boldsymbol{v}^{-}\right) 
= -\log \left[\frac{\exp \left(\boldsymbol{v} \cdot \boldsymbol{v}^{+} / \tau\right)}{\exp \left(\boldsymbol{v} \cdot \boldsymbol{v}^{+} / \tau\right)+\sum_{n=1}^N \exp \left(\boldsymbol{v} \cdot \boldsymbol{v}_n^{-} / \tau\right)}\right]
\label{equ:contrastive}
\end{align}
where $\boldsymbol{v}$, $\boldsymbol{v}^{+} \in \mathbb{R}^{D}$ is a positive pair, $\boldsymbol{v}^{-} \in \mathbb{R}^{N \times D}$ is $N$ negative samples, and $\tau$ is a temporal parameter.

To enforce this constraint, we propose a temporal shuffling method and a sampling strategy. The encoder $q_\phi^{enc}$ includes a temporally disentangled encoding stage $\mathcal{G}_l$ at each scale $l \in \{1, 2, \dots, L\}$. We select disentangled features at all scale levels from generated night events features $q_\phi^{enc}(X_1(X_{t_i}))$.
Given the result of the encoding stage $\mathcal{G}_l$ at scale $l$-th, $\tilde{F}_l = [\tilde{F}^1_l, \tilde{F}^2_l, \dots, \tilde{F}^B_l]$, to extract negative samples, we shuffle the temporal order of these features.
Using the permutations without a sorted (1, 2, ..., $B$) sample, we get ${{}_{B}\mathrm{P}_{B}}-1$ samples, \eg $\tilde{F}_{l,j} = [\tilde{F}^3_l, \tilde{F}^B_l, \dots, \tilde{F}^1_l]$, where $0 < j < \text{P}(B,B)$. 
We denote the feature stack with $R$ random selections from channel-wise shuffling as $\tilde{P}_l = \{\tilde{F}_{l,1}, \tilde{F}_{l,2}, \dots, \tilde{F}_{l,R} \}$.
We utilize an MLP network, $M_l$, to generate reference and negative features as $\{z_l\}_L = \{M_l(\tilde{F}_l)\}_L$ and $\{z_l^-\}_L = \{M_l(\tilde{P}_l)\}_L$, respectively. We select spatial positions $s \in {1, ..., S_l}$, with $S_l$ representing the number of spatial locations in each layer. We denote the corresponding features as $z_{l,s} \in \mathbb{R}^{C_l}$ and $z_{l,s}^{-} \in \mathbb{R}^{R \times C_l}$, where $C_l$ is the number of channels in each layer. Similarly, we transform the day event $X_0$ into $\{z^+_l\}_L = \{M_l(q_\phi^{enc}(X_0))\}_L$. Finally, the proposed temporally shuffling regularization can be calculated as follows:
\begin{align}
\mathfrak{L}_{TC}\left(\phi, t_i\right) = \mathbb{E}_{p(X_0, X_{t_i})} \mathbb{E}_{q_{\phi_{i}}(X_1|X_{t_i})}\sum_{l=1}^L \sum_{s=1}^{S_l} \mathfrak{L}_{C}(z_{l,s}, z_{l,s}^{+}, z_{l,s}^{-}).
\label{equ:contrastive}
\end{align}
\noindent
More information on temporally shuffling regularization, such as sampling strategy and pseudo-code, can be seen in supplementary material.

\section{Experiments}

\subsection{Setup}
\noindent
\textbf{Implementation Details.} We build our implementation based on the setting of USNB~\cite{kim2023unpaired} and CUT~\cite{park2020contrastive}.
In training steps, we tuned the U-Net-based generator~\cite{ronneberger2015u}.
We train our method by cropping and resizing events to 256$\times$256.

\noindent
\textbf{Dataset.}
DSEC~\cite{Gehrig21ral} dataset comprises event and image data depicting driving scenes from their sensor system. The dataset includes semantic segmentation labels, object bounding boxes, and other annotations.
We partitioned the existing DSEC dataset into day and night events based on criteria to achieve the goal.

\noindent
\textbf{Evaluation Metric.}
To the best of our knowledge, there has been no evaluation metric of the unpaired day-to-night event translation; hence, we adopt evaluation techniques of similar tasks, unpaired image-to-image translations~\cite{zhu2017unpaired, chen2020reusing, wang2022fregan}.
To evaluate for bin 1, we employ the FID score~\cite{heusel2017gans}. Where the bin size is greater than 1, we adopt the FVD score~\cite{unterthiner2018towards}, commonly used in video tasks, emphasizing the importance of temporal consistency.
Since existing metrics are tailored to an image modality, we train Inception-v3~\cite{szegedy2016rethinking} for FID and a 3D Convnet (I3D)~\cite{carreira2017quo} for FVD adapting to events.
Additionally, we convert translated events to an image and evaluate them.
E2I is conducted by E2VID~\cite{rebecq2019high} trained on night events and paired images. For E2I evaluation, we apply the FID and KID scores~\cite{bińkowski2018demystifying}. More details about the setup are described in the supplementary materials.

\begin{table}[t!b]
    \centering
    \caption{\textbf{Quantitative results} with FID, FVD scores for events and E2I-FID, E2I-KID$\times100$ scores for the E2I metric (For all metrics, lower is better). The best and second-best scores are highlighted and underlined. [ ] represent results generated from different sequences. v2e and ${\mathrm{v2e}}^{\dagger}$ were simulated with a 200 Hz cutoff frequency, while ${\mathrm{v2e}}^{\ast}$ used a 10 Hz cutoff to better represent nighttime conditions.}
    \resizebox{0.99\linewidth}{!}{
    \begin{tabular}{lcccccccccccc}
        \hline
        \multicolumn{1}{c}{} &  & \multicolumn{3}{c}{2 Channel (Bin = 1)} &  & \multicolumn{3}{c}{6 Channel (Bin = 3)} &  & \multicolumn{3}{c}{16 Channel (Bin = 8)} \\ \cline{3-5} \cline{7-9} \cline{11-13} 
        Method &  & \multicolumn{1}{c}{FID} & \multicolumn{1}{c}{E2I-FID} &E2I-KID &  & \multicolumn{1}{c}{FVD } & \multicolumn{1}{c}{E2I-FID } & \multicolumn{1}{c}{E2I-KID } &  & \multicolumn{1}{c}{FVD } & \multicolumn{1}{c}{E2I-FID } & \multicolumn{1}{c}{E2I-KID } \\ \hline \hline
        Original Day Event & & 1280.15 & 45.56 & 4.78 & & 332.18 & 45.81 & 4.92 & & 134.68 & 46.43 & 5.35 \\ \hline
        Simulator & & & & & & & & & & & &  \\ \hline
        v2e~\cite{hu2021v2e} (using day img.) &  & 950.16 & 52.63 & 5.87 &  & 47.05 & 37.93 & 5.61 &  & 22.42 & 49.50 & 9.05 \\
        ${\mathrm{v2e}}^{\ast}$~\cite{hu2021v2e} (using day img.)&  & 745.38 & 129.86 & 14.79 &  & 52.51 & 109.76 & 17.10 &  & 28.57 & 156.46 & 23.63 \\
        ${\mathrm{v2e}}^{\dagger}$~\cite{hu2021v2e} (using night img.)&  & [992.44] & [71.18] & [7.46] &  & [47.14] & [46.72] & [6.74] &  & [28.61] & [75.70] & [11.72] \\
        DVS-Voltmeter~\cite{lin2022dvs} (using day img.) &  & 1905.87 & 57.15 & 5.94 &  & 276.75 & 51.54 & 6.31 &  & 193.21 & 59.47 & 7.25 \\ 
        DVS-${\mathrm{Voltmeter}}^{\ast}$~\cite{lin2022dvs} (using night img.)&  & [888.75] & [30.07] & [\underline{3.90}] &  & [158.96] & [33.65] & [4.79] &  & [113.29] & [44.29] & [6.80] \\ \hline
        Frame & & & & & & & & & & & &  \\ \hline
        EventGAN~\cite{zhu2021eventgan} (trained on day img.) &  & 1270.95 & 136.74 & 16.06 &  & 135.90 & 109.45 & 23.56 &  & 34.41 & 99.99 & 18.62 \\
        ${\mathrm{EventGAN}}^{\ast}$~\cite{zhu2021eventgan} (trained on night img.)&  & 649.33 & 113.11 & 18.45 &  & 29.86 & 109.29 & 10.53 &  & 54.80 & 129.74 & 20.57 \\
        CMDA~\cite{xia2023cmda} &  & 1068.79 & 31.11 & 5.03 &  & 49.49 & 82.42 & 10.64 &  & 20.83 & 115.19 & 25.70 \\ \hline
        Event & & & & & & & & & & & &  \\ \hline
        CycleGAN~\cite{zhu2017unpaired} &  & 1670.76 & 27.29 & 3.96 &  & 57.11 & 33.22 & 4.87 &  & 39.17 & 74.84 & 5.64 \\
        GcGAN~\cite{fu2019geometry} &  & 658.62 & 43.80 & 5.58 &  & \underline{44.98} & 29.48 & 3.85 &  & 42.11 & 30.24 & 4.70 \\
        CUT~\cite{park2020contrastive} &  & 1462.74 & 28.23 & 4.86 &  & 58.90 & \underline{25.66} & \underline{3.21} &  & \textbf{9.72} & 28.58 & 3.43 \\
        UNSB~\cite{kim2023unpaired} &  & \textbf{352.58} & \underline{26.87} & 5.38 &  & 74.25 & 26.74 & 4.48 &  & 34.38 & \underline{25.05} & \underline{3.31} \\
        \textbf{Ours} &  & \underline{376.21} & \textbf{23.20} & \textbf{3.74} &  & \textbf{7.19} & \textbf{18.01} & \textbf{1.24} &  & \underline{10.14} & \textbf{23.09} & \textbf{1.77} \\ \hline
    \end{tabular}}
    \label{tab:main}
\end{table}

\subsection{Quantitative Results}

In~\cref{tab:main}, we quantitatively compare our method with event simulators (v2e~\cite{hu2021v2e}, DVS-Voltmeter~\cite{lin2022dvs}), image-based translation methods (EventGAN~\cite{zhu2021eventgan}, CMDA \cite{xia2023cmda}), and event-based translation methods (CycleGAN~\cite{zhu2017unpaired}, GcGAN~\cite{fu2019geometry}, CUT \cite{park2020contrastive}, UNSB~\cite{kim2023unpaired}.)
For the 2-channel, our key modules, TDE and $\mathfrak{L}_{TC}$, do not exist, resulting in a small gap compared to other methods. For 6-channel cases, our proposed temporal regularization effectively considers the properties of events, leading to a significant performance gap compared to others. However, the challenge with the 16-channel is that events are divided into 8 time intervals, resulting in very sparse events.
Therefore, while the gap may decrease for the style-focused FVD, we still observe a significant margin over others for E2I-KID, which considers the temporal properties of events. Based on the results, our approach successfully produces night events from day domain data across all channel cases.

\begin{figure}[t]
    \centering
    \includegraphics[width=0.99\linewidth]{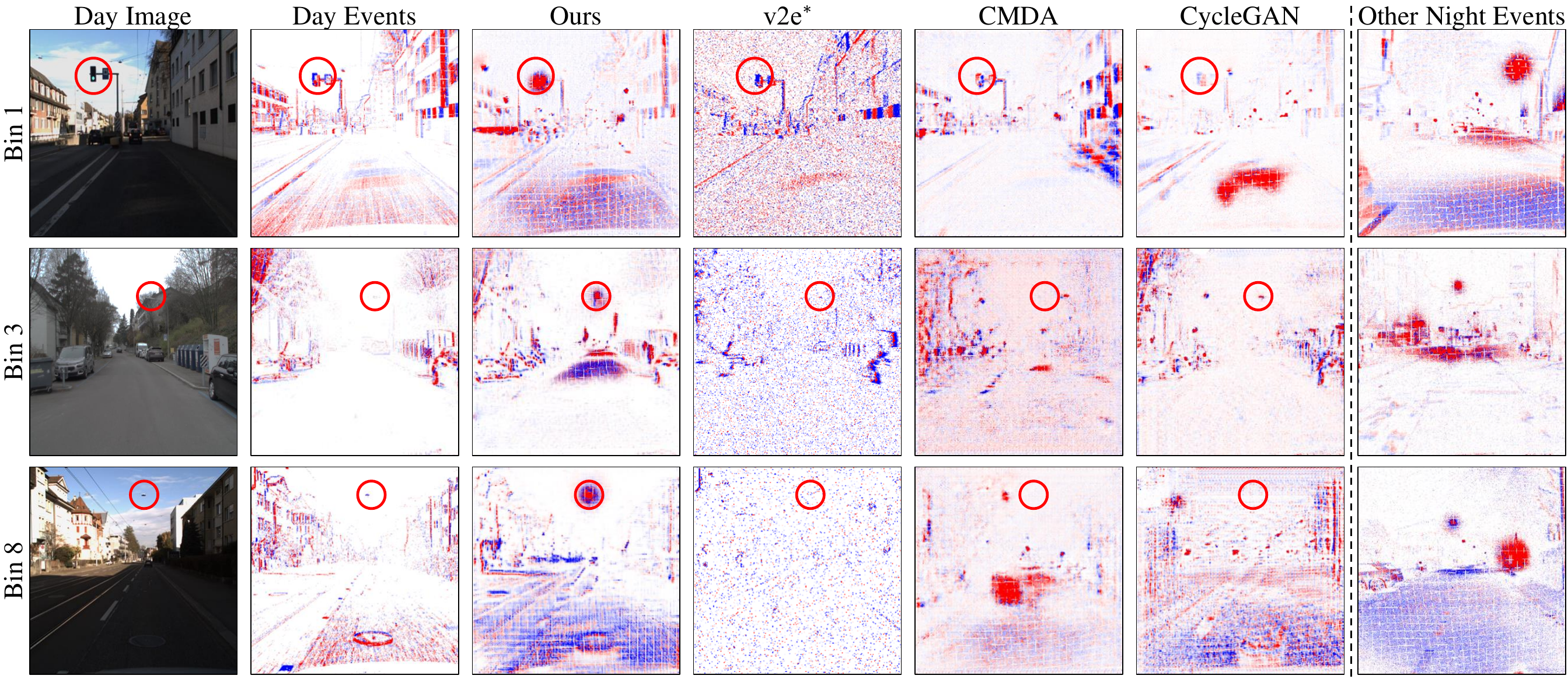}
    \caption{\textbf{Qualitative results} of translated night events of each method. The red and blue dots mean positive and negative polarity events, respectively. The first column serves as an image aligned with input day events in the second column. Other columns show the results of each method. The last column visualizes reference night events.}
    \label{fig:qual_1}
\end{figure}

\subsection{Qualitative Results}
The qualitative comparison between our approach and prior research is depicted in Fig.~\ref{fig:qual_1}.
Simulators effectively generate the natural noise of night events.
However, the simulator struggles to reproduce another distinctive feature of night events related to illuminants.
Image-based generation methods preserve the structure but tend to exhibit different patterns from real events. These methods, originating from intensity changes in sequential images, cannot effectively produce outputs for a high dynamic range camera.
Event-based translation methods fail to generate the natural noise of night events and fill the background with meaningless values, resulting in unrealistic events.
In contrast, our approach aligns more closely with actual event characteristics. Our method effectively learns the distinct properties of night events through wavelet decomposition to reproduce the night style. All methods utilize grid-based events during training, generating grid-like artifacts. However, this is not a methodological issue, and these artifacts can be removed if rectification is not applied during the training.

\subsection{Ablation Study}
\noindent
\textbf{Effectiveness of each component of our framework.}
We design several ablation experiments by adding and removing the proposed module in~\cref{tab:main_ablation}. The baseline (a) is the framework only with the spatial branch and the spatial contrastive regularization.
In (b), the spatial branch is replaced with the frequency branch in the mid block.
This entails a performance trade-off.
The transfer of insufficient contextual knowledge worsens the FVD, whereas the frequency branch adeptly delivers the style of night events, elevating the E2I-FID.
Using the spatial and frequency branches together (c) does not lead to the optimal path.
The worsened FVD score, which measures events' temporal properties, indicates the necessity to enhance their temporal consistency.
When we disentangle the encoder, the improved E2I-FID scores are achieved (\eg,~from (a) to (d): 41.81 $\rightarrow$ 18.58).
The temporally disentangled encoder convolutes features within each bin to preserve their temporal continuity,
effectively generating the style of the night.
However, disentanglement weakened the sharing of structural information, resulting in a loss of structure.
As a consequence, the FVD score deteriorates (\eg,~from (b) to (f), 13.64 $\rightarrow$ 44.87).
The $\mathfrak{L}_{TC}$, aimed at enhancing the temporal continuity of events.
Overall, applying $\mathfrak{L}_{TC}$ improves all metrics except for a slight increase in E2I-KID from (d) to (e). This indicates that the $\mathfrak{L}_{TC}$ not only strengthens temporal continuity but also maintains the structural coherence of the event, going beyond the mere enhancement of temporal aspects. Using all the proposed modules simultaneously yielded the best results. 

\begin{table}[t!]
    \centering
    \caption{\textbf{Ablation} on our proposed modules. Lower is better.}
    \setlength\tabcolsep{6.2pt}
    \resizebox{.94\linewidth}{!}{
    \begin{tabular}{ccccccccl|ccc}
        \hline
        & Spatial Branch & & Frequency Branch & & TDE & & $\mathfrak{L}_{TC}$ &  & FVD & E2I-FID & E2I-KID \\ \hline
        (a) & \checkmark & & & & &  &  &  & 9.92 & 41.81 & 5.57 \\
        (b) &  & & \checkmark & & & &  &  & 13.64 & 39.25 & 5.56\\
        (c) & \checkmark & & \checkmark & & & &  & & 17.70 & 24.05 & 3.29 \\
        (d) & \checkmark & & & & \checkmark & & &  & 9.89 & 18.58 & 1.88 \\
        (e) & \checkmark & & & & \checkmark & & \checkmark &  & 8.85 & 18.05 & 2.01 \\
        (f) &  & & \checkmark & & \checkmark & & &  & 44.87 & 30.59 & 5.43 \\
        (g) & & & \checkmark & & \checkmark & & \checkmark &  & 16.92 & 21.00 & 2.18 \\
        (h) & \checkmark & & \checkmark & &\checkmark & & &  & 8.98 & 21.31 & 1.95 \\
        (i) & \checkmark & & \checkmark & & \checkmark & & \checkmark &  & \textbf{7.19} & \textbf{18.01} & \textbf{1.24} \\ \hline
    \end{tabular}}
    \label{tab:main_ablation}
\end{table}

\begin{figure}[t]
    \centering
    \includegraphics[width=.94\linewidth]{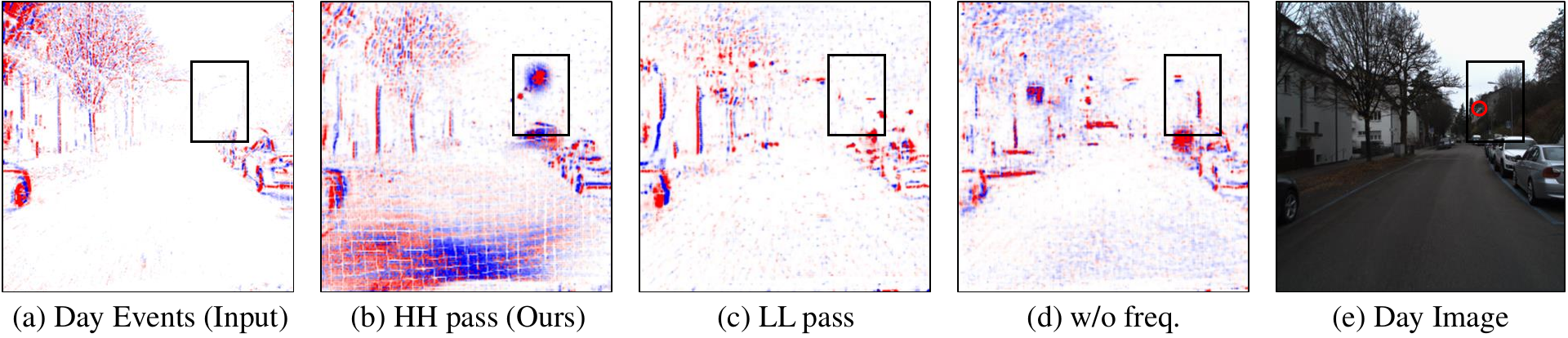}
    \caption{Visual comparisons of applying wavelet decomposition in the frequency branch. (a) shows the input day events. (b) and (c) show the results of applying the frequency branch passing HH and LL components through ResNet, respectively. (d) shows the result without the frequency branch. (e) shows an aligned image of the input.}
    \label{fig:wavelet_anlysis2}
\end{figure}

\begin{figure}[t]
    \centering
    \includegraphics[width=0.95\linewidth]{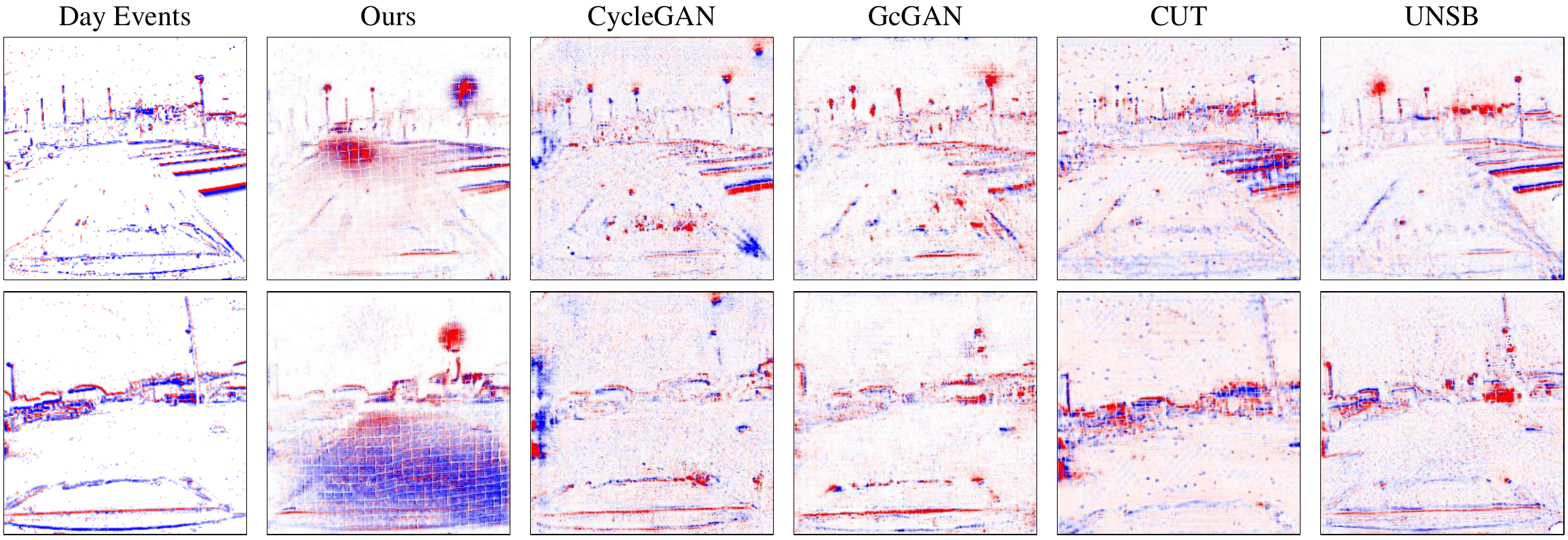}
    \caption{Comparisons in unseen MVSEC~\cite{zhu2018multivehicle} using the model trained on DSEC~\cite{Gehrig21ral}.}
    \label{fig:mvsec}
\end{figure}

\noindent
\textbf{Applying Wavelet Decomposition to events.}
We conducted an additional analysis to delve more deeply into the effects of the frequency branch utilizing wavelet decomposition. In~\cref{tab:wavelet_analysis2}, we set the baseline for this experiment (e) of the ablation study in~\cref{tab:main_ablation}. Furthermore, we compare the results by processing the LL component instead of the HH component in the frequency branch. The frequency domain information enhances the structure of generated events, with any frequency branch addition improving the FVD score.
The addition of any form of frequency branch achieves better FVD.
However, unlike HH, passing through the LL component leads to worse E2I-FID/KID. This indicates the inability to implement the night style in the structure while manipulating the information from the LL component. It can also be seen in Fig~\ref{fig:wavelet_anlysis2}, where, despite effectively reproducing the structure, the module that processes the LL component fails to generate a style resembling night.

\begin{table}[t]
    \begin{minipage}[t]{0.47\linewidth}
        \centering
        \captionof{table}{Experiments on replacing the HH component with the LL component in the frequency branch.}
        \resizebox{0.99\linewidth}{!}{
        \begin{tabular}{llccc}
            \hline
            Method &  & FVD & E2I-FID & E2I-KID \\ \hline
            w/o Freq. &  & 8.85 & \underline{18.05}  & \underline{2.01} \\ 
            LL Freq. pass &  & \textbf{7.17} & 22.24 & 3.03 \\ 
            HH Freq. pass (Ours) &  & \underline{7.19} & \textbf{18.01} & \textbf{1.24} \\\hline
        \end{tabular}}
        \label{tab:wavelet_analysis2}
    \end{minipage}
    \hfill
    \begin{minipage}[t]{.475\linewidth}
        \centering
        \setlength\tabcolsep{6.5pt}
        \captionof{table}{Comparison in unseen MVSEC \cite{zhu2018multivehicle} environments using our method and others trained on DSEC~\cite{Gehrig21ral}.}
        \resizebox{0.87\linewidth}{!}{
        \begin{tabular}{lccc}
            \hline
            Method & CycleGAN~\cite{zhu2017unpaired} & CUT~\cite{park2020contrastive} \\ \hline
            FVD & 294.94 & \underline{158.39} \\ \hline
            Method & UNSB~\cite{kim2023unpaired} & Ours \\ \hline
            FVD & 322.60 & \textbf{130.47} \\ \hline
        \end{tabular}}
        \label{tab:mvsec}
    \end{minipage}
\end{table}

\subsection{Validation in the Unseen Dataset}
We apply our proposed method in an unseen scenario to validate that our model is generalizable for various datasets. We choose MVSEC~\cite{zhu2018multivehicle}, which also contains day and night outdoor scenarios, and do experiments with the model trained on DSEC~\cite{Gehrig21ral}. For the MVSEC dataset, we split events into the same representation of bin 3. The comparison with translation methods can be seen in Tab.~\ref{tab:mvsec} and Fig.~\ref{fig:mvsec}. The result shows that our proposed method showcases performance stability even with datasets not seen during the training.

\subsection{Applications on Downstream Tasks}

We train the network using both day events and translated night events as following (c) of Fig.~\ref{fig:intro_night_model}. We set the network trained only with day events as a baseline for all downstream tasks. We compare the performance of several downstream tasks while changing the translation methods, inferred with real night events, in Fig.~\ref{fig:application}. The details of each task are described in the supplementary material.

\begin{figure}[t]
    \centering
    \includegraphics[width=0.95\linewidth]{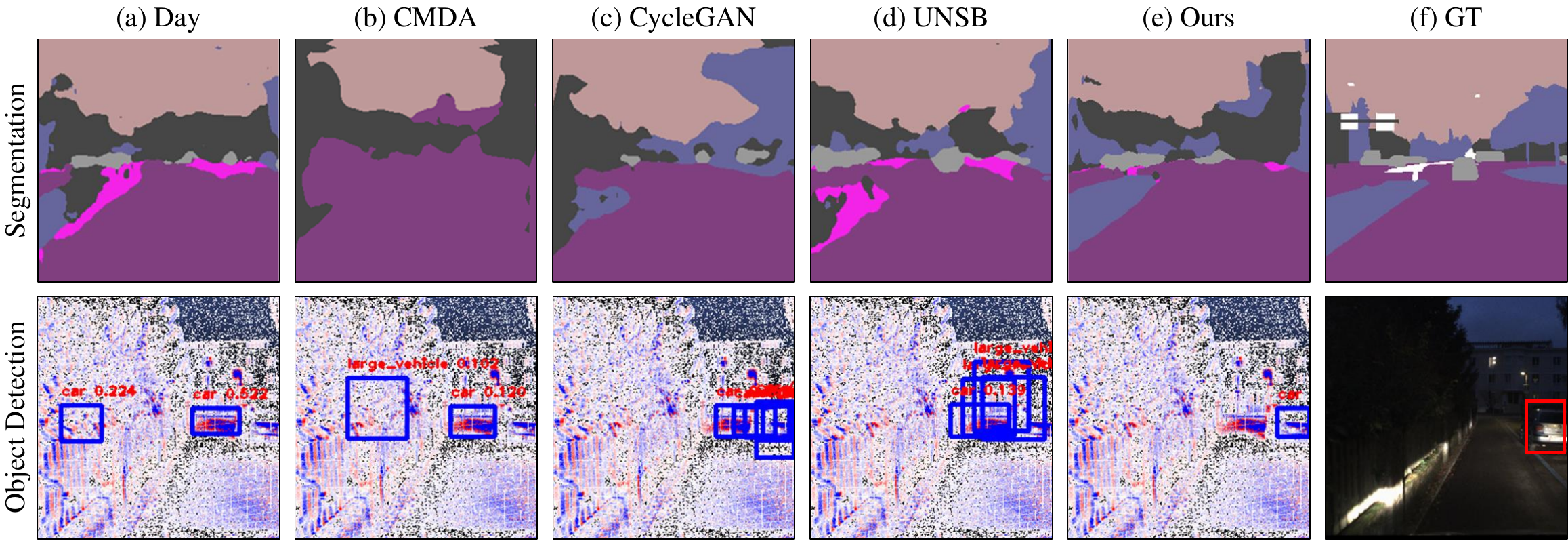}
    \caption{Visual comparison of downstream tasks trained with day events and translated events, inferred with real night events. (a) is trained with day events only.}
    \label{fig:application}
\end{figure}

\noindent
\textbf{Nighttime Event-based Semantic Segmentation.}
We present the results of a segmentation task in~\cref{tab:segmentation}. Most methods, except EventGAN, contribute to overall performance improvement in bin 1. Image-based methods struggle with multi-bins due to their inability to consider the temporal continuity of events. Event-based methods, which maintain continuity, outperform the baseline for bin 3. In the case of bin 8, all methods except ours decrease performance compared to baseline, showing the lack of approaches to temporal continuity.

\noindent
\textbf{Event-based Object Detection.}
We compare the object detection results in~\cref{tab:ob}. Most methods, except CMDA, lead to an overall improvement in bin 1. However, for bin 3, all methods, except ours, could not avoid setbacks.
Mainly, EventGAN, an image-based generation method, exhibits the most significant decline for multi-bins.
In contrast, our method performs better than the baseline.

\begin{table}[t!]
    \begin{minipage}[t]{.49\linewidth}
        \centering
        \captionof{table}{The results of night semantic segmentation training with translated night events. mIoU is adopted for the metric.}
        \renewcommand{\arraystretch}{1.1}
        \resizebox{0.99\linewidth}{!}{
            \begin{tabular}{l|cccc}
                \hline
                Method &  Bin 1 &  Bin 3 &  Bin 8 \\ \hline
                Day Event & 26.99 & 34.28 & \underline{36.65}\\
                ${\mathrm{EventGAN}}^{\ast}$~\cite{zhu2021eventgan} (trained on Night) & 18.96 & 28.45 & 25.43 \\
                CMDA~\cite{xia2023cmda}  & 36.44  & 27.28 & 23.91\\ 
                CycleGAN~\cite{zhu2017unpaired} &  33.25 & 38.92 & 25.04 \\
                UNSB~\cite{kim2023unpaired} & \underline{36.45} & \underline{38.98} & 34.76 \\
                \textbf{Ours} & \textbf{37.80} & \textbf{39.49} & \textbf{39.58} \\ \hline
            \end{tabular}}
        \label{tab:segmentation}
        
    \end{minipage}
    \hfill
    \begin{minipage}[t]{.48\linewidth}
        \centering
        \captionof{table}{The results of night object detection training with translated night events. AP is adopted for the metric.}
        \resizebox{0.99\linewidth}{!}{
            \begin{tabular}{l|cc}
            \hline
            Method & Bin 1 & Bin 3  \\ \hline
            Day Event & 3.93 & \underline{8.87}  \\
            ${\mathrm{EventGAN}}^{\ast}$~\cite{zhu2021eventgan} (trained on Night) & \underline{6.97} & 3.21   \\
            CMDA~\cite{xia2023cmda} & 3.03 & 4.78  \\
            CycleGAN~\cite{zhu2017unpaired} & 5.84 & 8.39  \\
            UNSB~\cite{kim2023unpaired} & 5.38 & 4.25  \\
            \textbf{Ours} & \textbf{9.65} & \textbf{13.55}  \\ \hline
            \end{tabular}}
        \label{tab:ob}
    \end{minipage}
\end{table}

\section{Conclusion, Limitations and Future Work}
In this paper, we propose a framework to solve the challenging task of translating unpaired day-to-night events. We preserve the characteristics of events during translation through a novel temporally shuffling contrastive regularization with a disentangled encoder and wavelet transform. Additionally, we introduce metrics to evaluate events quantitatively.
By translating annotated day events into night events and using them to train event-based models, we improve performance when applying real night events to these models. The applications show the effectiveness and task-agnostic use cases of our methods.
Our work has a limitation: most generation models rely on convolutional layers, requiring the conversion of events into frame-wise representations for utilization. 
For similar reasons, many event-based networks~\cite{liu2018adaptive, duan2021eventzoom, weng2022boosting, gehrig2021raft, xia2023cmda, sun2022event, liu2023tma} convert events into voxel or frame formats for utilization. By developing models using SNNs~\cite{yao2021temporal, zhu2022event, hagenaars2021self, lee2020spike, zhu2021neuspike} or graph-based networks~\cite{li2021graph, schaefer2022aegnn, bi2019graph}, we can improve our framework for natural events. Moreover, our framework is fundamentally optimized for autonomous driving data and has demonstrated the conversion of event data on the road scenes. However, we have yet to explore the conversion of general events between day and night. We plan to address this in future research.

\noindent
\textbf{Acknowledgements.} This work was supported by the National Research Foundation of Korea(NRF) grant funded by the Korea government(MSIT) (NRF2022\\R1A2B5B03002636).


%
%

\bibliographystyle{splncs04}
\bibliography{egbib}

\renewcommand{\thetable}{\Alph{table}}
\renewcommand{\thefigure}{\Alph{figure}}


%


\definecolor{red}{rgb}{1, 0, 0}
\definecolor{blue}{rgb}{0, 0, 1}

\title{Towards Robust Event-based Networks for Nighttime via Unpaired Day-to-Night Event Translation \\ \centering{\textit{---Supplementary Material---}}}

\titlerunning{Towards Robust Event-based Networks for}

\author{Yuhwan Jeong\orcidlink{0009-0002-0279-146X}\thanks{Equal contribution.} \and
Hoonhee Cho\orcidlink{0000-0003-0896-6793}\protect\CoAuthorMark \and
Kuk-Jin Yoon\orcidlink{0000-0002-1634-2756}}

\authorrunning{Jeong et al.}

\institute{Korea Advanced Institute of Science and Technology\\
\email{\{jeongyh98, gnsgnsgml, kjyoon\}@kaist.ac.kr}\\
}

\newpage

\maketitle

\appendix

Our supplementary material provides additional information about our proposed method and includes detailed discussions on the following contents that were not extensively covered in the main paper:

\begin{itemize}
    \item Details about the unpaired translation using Schrödinger Bridge (Sec.~\ref{sec:preliminary}).
    \item Pseudocode for temporally shuffling contrastive regularization (Sec.~\ref{sec:pseudo_code}).
    \item More details about the implementation, architecture, and dataset (Sec.~\ref{sec:experiment_detail}).
    \item Details about the redesigned evaluation metrics and analyses for metrics (Sec.~\ref{sec:metric}).
    \item Additional analysis and discussion of the proposed method (Sec.~\ref{sec:discussion}).
    \item More qualitative results of day-to-night event translation (Sec.~\ref{sec:more_qual}).
    \item Details of applications on downstream tasks that utilize translated night events from annotated day events (Sec.~\ref{sec:application}).
    \item Opposite direction: night-to-day event translation (Sec.~\ref{sec:night-to-day}).
\end{itemize}




\section{Unpaired Image-to-Image Translation via Neural Schrödinger Bridge.} 
\label{sec:preliminary}
Paired translation changes style when structures are identical. On the other hand, in an unpaired setting, neither the style nor the structure of day and night events is identical. Therefore, the key challenge is to maintain the structure while only changing the style, which often leads to divergence and collapse. Event modality exacerbates these issues compared to images due to the additional temporal dimension. To address these issues, various regularizations are required. 

In this section, we provide more detailed information about the unpaired translation via Schrödinger Bridge~\cite{kim2023unpaired}, which combines score-based diffusion and discriminator. Schrodinger Bridge(SB) addresses optimal transport (OT) for unpaired translation using entropy regularization and bidirectional processes between distributions. In contrast, other OT methods (e.g., Wasserstein distance and maximum mean discrepancy), which are non-dynamic processes, often produce sparse plans, compromising translation quality. These two features of SB prevent overfitting and mode collapse, which are common in complex data distributions~\cite{wang2021deep}. We also attempted to explore alternative OT approaches, but the SB process consistently 
modeled more robust solutions for unpaired translation. The following section covers extra materials not extensively covered in the preliminary section of the main paper.

In our problem, we have two distributions, which correspond to $E_D$ and $E_N$, denoted as \(\pi_0\) and \(\pi_1\)  on $\mathbb{R}^d$.
We aim to find the optimal transport path from \(\pi_0\) to \(\pi_1\) using the Schrödinger Bridge problem (SBP)~\cite{de2021diffusion}.
The SBP is to find the most likely random stochastic process that connects two distributions. More specifically, the SBP explores the random process $\{X_t :t \in [0, 1]\}$ which follows the Schrödinger bridge, $\mathbb{Q}^{*}$, between day and night event distributions $\pi_0, \pi_1$. It can be solved by:
\begin{equation}
    \mathbb{Q}^{*} = \underset{\mathbb{Q} \in P(\Omega)}{\mathrm{argmin}} \mathbb{D}_{KL}(\mathbb{Q}\parallel \mathbb{W}_{\tau}),~\text{s.t.}~\mathbb{Q}_0 = \pi_0, \mathbb{Q}_1=\pi_1
    \label{eqn:Q_*}
\end{equation}
where $\mathbb{W}_{\tau}$ is the Wiener measure with variance $\tau$, $\Omega$ is the space of continuous functions on time interval $t$, and $P(\Omega)$ denotes the space of probability measure on $\Omega$. 
We can also set the marginal of path $\mathbb{Q}^{*}$ at time $t$ as  $\mathbb{Q}_{t}^{*} = \mathbb{Q}^{*} \circ X_t^{-1}$. 
To build the random process, $\left\{ X_{t}\right\}$ $\sim$ $\mathbb{Q}^{*}$, with a stochastic step, we use the stochastic control formulation~\cite{dai1991stochastic} in the form of SDE as follows:
\begin{align}
\mathrm{d}{X_t} = \textbf{u}_{t}^{*} \, \mathrm{d}t + \sqrt{\tau} \, \mathrm{d}w_{t}.
\label{eqn:dX}
\end{align}
where $w$ is the Wiener process. 
A time-varying drift vector, $\textbf{u}_{t}^{*}$, solves the SDE when $\mathfrak{U}$ satisfies Markov property. 
as:

\begin{align}
    \textbf{u}_{t}^{*}  =  \underset{u \in \mathfrak{U}}{\operatorname{argmin}} \, \mathbb{E} \left [\int_{0}^{1}\frac{1}{2}\left\| \textbf{u}_{t}\right\|^{2}\mathrm{d}t \right] \\
     \text{s.t.} \left\{\begin{matrix}
    \mathrm{d}{X_t} = \textbf{u}_{t} \mathrm{d}t + \sqrt{\tau} \, \mathrm{d}w_{t} \\ X_{0} \sim \pi_{0},\quad  X_{1} \sim \pi_{1}
    \end{matrix}\right.
    \label{eqn:vector_field}
\end{align}

According to Schrödinger Bridge continuous flow matching (CFM)~\cite{tong2023improving}, while the initial condition $X_{0}$ is given, the conditional probability can plan the marginal process $X_{t}$.
As converge \(X_t\) to the optimal path, Schrödinger bridge CFM, which is an entropic variant of optimal transport CFM, simply describes the path $\mathbb{Q}^{*}$ in the viewpoint of ODE. As terminal conditions are given, a probability flow $p(X_t)$ can be described by:
\begin{align}
p(X_t|z)= \mathit{N}(X_t|tX_1+(1-t)X_0, t(1-t)\tau I),
\label{eqn:probability_flow}
\end{align}
where $z$ is a joint distributions of the terminal conditions, $X_{0}$ and $X_{1}$. While the initial condition $X_{0}$ is given, the conditional probability,
\begin{align}
p(X_t|X_0)=\int p(X_t | z) \mathrm{d} \, \mathbb{Q}_{1|0}^{*}(X_1|X_0),
\label{eqn:sampling}
\end{align}
can plan the marginal process $X_{t}$.
Considering the entropy-regularized optimal transport problem~\cite{hernandez2012discrete} with two distributions $\pi_0$ and $\pi_1$, Schrödinger Bridge problem should be solved by:
\begin{align}
\mathbb{Q}_{01}^{*}=\underset{{\pi \in\Pi(\pi_0,\pi_1)}}{\operatorname{argmin}}\mathbb{E}_{(X_0,X_1) \sim \pi}\left [ {\left\| X_0 - X_1 \right\|}^2 \right ]-2\tau H(\pi),
\label{eqn:final}
\end{align}
where $H$ denotes the entropy function of transport distribution $\pi$.
According to Sinkhorn algorithm, for two finite discrete distributions, the optimal transport can be planed, and we can set the discrete intervals for approximated SBP~\cite{genevay2019entropy}.

As following ~\cite{kim2023unpaired}, we set the interval $[t_a, t_b] \subseteq [0, 1]$ to restrict the SB, optimal transport plan becomes

\begin{align}
    \mathbb{Q}_{t_a t_b}^{*}=\underset{\pi \in\Pi(Q_{t_a},Q_{t_b})}{\operatorname{argmin}}\mathbb{E}_{(X_{t_a},X_{t_b}) \sim \pi} \left[{\left\| X_{t_a} - X_{t_b} \right\|}^2 \right ]-2\tau(t_b-t_a)\, H(\pi).
    \label{eqn:res_final}
\end{align}
Using a Markov chain from $t_0 = 0$ to $t_M = 1$, we unfold the probability flow.
We formulate the path of two distributions recursively as follows:
\begin{align}
    p(\{X_{t_{m}} \}) = p(X_{t_{M}}|X_{t_{M-1}})\cdots \, p(X_{t_{1}}|X_{t_{0}}) p(X_{t_{0}}).
    \label{eqn:flow_markov}
\end{align}

To seek the SB process $p$, we define $q_{\phi_i}$ as a conditional distribution
parameterized by a DNN with parameter $\phi_i$. Then, to stabilize our path as the optimal path and to optimize $q_{\phi_i}$, we utilize the below losses while applying restricted interval when $t_a=$ and $t_b=1$: 

\begin{align}
    \underset{\phi_i}{\textnormal{min}}\;\mathfrak{L}_{*}(\phi_i, t_i) = \mathbb{E}_{q_{\phi_{i}}(X_{t_i}, X_1)} \left [ \left\| X_{t_i}-X_1\right\|^2\right ] -2\tau(1-t_i)H(\phi_i (X_{t_i}, X_1))
    \label{eqn:final_const1}
\end{align}

\begin{align}
    \textnormal{s.t.}\quad\mathfrak{L}_{\textnormal{Adv}}(\phi_i, t_i)=D_{\textnormal{KL}}(q_{\phi_i}(X_1)\parallel p(X_1)) = 0. 
    \label{eqn:final_const22}
\end{align}

\section{Pseudocode for Temporally Shuffling Contrastive Regularization}
\label{sec:pseudo_code}
We provide the pseudo-code of temporally shuffling contrastive regularization loss, $\mathfrak{L}_{TC}$ in Algorithm~\ref{algo:pseudo_code_tsr}. We implemented this regularization using PyTorch and achieved efficiency by performing most operations as batch operations.

\begin{algorithm}[ht!]
    \caption{Temporally Shuffling Contrastive Regularization}
    \label{algo:pseudo_code_tsr}
    \textbf{Input:} \\
    Encoded features, $q_\phi^{enc}(X_0))$, from day events\\
    Encoded features, $q_\phi^{enc}(X_1(X_{t_i}))$, from translated night events\\
    \textbf{Output:} \\
    Temporally Shuffling Contrastive Regularization Loss, $\mathfrak{L}_{TC}$ \\
    \textbf{Algorithm:}
    \begin{algorithmic}
        \State{\textcolor{blue}{$\#$ Initialize Loss.}}
        \State{$\mathfrak{L}_{TC}$  = 0}
        \For{$l$ = 1, \dots, $L$}
            \State{$\tilde{F}_l = [\tilde{F}^1_l, \tilde{F}^2_l, \dots, \tilde{F}^B_l]$ from $\mathcal{G}_l$ in $q_\phi^{enc}(X_1(X_{t_i}))$.}
            \State{\textcolor{blue}{$\#$ Calculate permutations excluding $(1,2,\dots,B)$}}
            \State{$\mathcal{P}$ $\leftarrow$  Permutation($\tilde{F}_l$)}~ s.t. ~Len($\mathcal{P}$) = $\text{P}(B,B)-1$.
            \State{$\mathcal{P} = \{\tilde{F}_{l,1}, \cdots, \tilde{F}_{l,j}, ,\cdots, \tilde{F}_{l, \text{P}(B,B)-1}\}$} such that
            \State{$\tilde{F}_{l,j} = [\tilde{F}^{j_1}_l, \tilde{F}^{j_2}_l, \dots, \tilde{F}^{j_B}_l]$ where $j_1 \neq j_2 \dots \neq j_B$}
            \State{and $\tilde{F}_{l,1} \neq \tilde{F}_{l,2} \neq \dots \neq \tilde{F}_{l,j} \neq \tilde{F}_{l}$.}
            \State{\textcolor{blue}{$\#$ Randomly choose $R$ samples from $\mathcal{P}$.}}
            \State{$\tilde{P}_l = \{\tilde{F}_{l,1}, \tilde{F}_{l,2}, \dots, \tilde{F}_{l,R} \}$.}
            \State{\textcolor{blue}{$\#$ Sample negative features using MLP, $M_l$.}}
            \State{$z_l^{-}$ = $M_l(\tilde{P}_l)$}
            \State{\textcolor{blue}{$\#$ Sample reference features using shared MLP, $M_l$.}}
            \State{$z_l$ = $M_l(\tilde{F}_l)$}
            \State{\textcolor{blue}{$\#$ Sample positive features using shared MLP, $M_l$.}}
            \State{$z_l^{+}$  = $M_l(\tilde{F}_l^{+})$ where  $\tilde{F}_l^{+}$ from $\mathcal{G}_l$ in $q_\phi^{enc}(X_0)$.}
            \State{Sample the same spatial positions, $s \in \{1, \dots, S_l\}$}.
            \State{$Z_l = \{z_{l,1}, \dots, z_{l,S_l}\}$, $Z_l^{+} = \{z_{l,1}^{+}, \dots, z_{l,S_l}^{+}\}$, and}
            \State{$Z_l^{-} = \{z_{l,1}^{-}, \dots, z_{l,S_l}^{-}\}$, where}
            \State{$Z_{l}, Z_{l}^{+} \in \mathbb{R}^{C_l \times S_l}$ and $Z_{l}^{-} \in \mathbb{R}^{C_l \times S_l \times R}$.}
            \State{\textcolor{blue}{$\#$ Calculate contrastive loss.}}
            \State{p\_logit = $(Z_l * Z_l^{+})$.sum(0).expand(1)} \Comment{$\mathbb{R}^{S_l \times 1}$}
            \State{n\_logit = $(Z_l\text{.expand(2)} * Z_l^{-})$.sum(0)} \Comment{$\mathbb{R}^{S_l \times R}$}
            \State{logit = concat(p\_logit, n\_logit)} / $\tau$ \Comment{$\mathbb{R}^{S_l \times (R + 1)}$}
            \State{loss = cross\_entropy\_loss(logit, 0)}
            \State{$\mathfrak{L}_{TC}$  += loss}
        \EndFor
    \end{algorithmic}
    \textbf{return} $\mathfrak{L}_{TC}$
\end{algorithm}

\section{More details about Experiment Setup}
\label{sec:experiment_detail}
\subsection{Implementation Details}
\label{sec:imple_details}
We set the value of $R$ in temporally shuffling contrastive regularization to 5 and 20 for bin 3 and bin 8, respectively.
In the case of bin 1, temporally shuffling contrastive regularization $\mathfrak{L}_{TC}$ is not applied. We train the network with a batch size of 1. For the temperature of spatial contrastive regularization, we followed the value of CUT~\cite{park2020contrastive}.
The temperature of our proposed temporally shuffling contrastive regularization, $\tau$, is 0.11.
Details of the temperature selection are described in Sec.~\ref{sec:temperature}. We used iterative paths with static timestamps $M$ as 5 and a hyperparameter $\lambda_{SB}$ as 1, the same as UNSB~\cite{kim2023unpaired}. For spatial contrastive regularization loss coefficient $\lambda_{SC}$, we followed a strategy of CUT~\cite{park2020contrastive}. For $\lambda_{TC}$, we set 1 for all cases.

We implemented a U-Net-like architecture based on the existing diffusion model~\cite{ho2020denoising} with LSGAN~\cite{mao2017least} strategy and PatchGAN~\cite{isola2017image} discriminator to our framework. More specifically, we utilized the publicly available pytorch code at the following link\footnote{\url{https://github.com/lucidrains/denoising-diffusion-pytorch/tree/main}}. One difference is that, in the upscaling process of the U-Net, we use nn.ConvTranspose2d() instead of nn.Upsample(). Empirically, we found this highly effective, especially in the event modality. We set $L$ as 5.
Moreover, there is no downsampling layer for the encoder's first and last scale.
Consequently, the event feature is downsampled 3 times.
In our approach, we removed the attention block in the encoder of general U-Net architecture to build the temporally disentangled encoder. 

As the number of bins increases, so does the spatio-temporal information of the events. To maintain temporal consistency amidst this progression, a high-dimensional feature embedding becomes required. Consequently, the channel dimension of the network was adjusted in correspondence with the increasing bin size. Specifically, for bin sizes of 1, 3, and 8, the channel dimensions of features were set to 64, 72, and 96, respectively. Similarly, the channel dimension of other approaches is set in the same way as ours.

\subsection{Implementation Details about Other Methods}

To compare our proposed method to others, we bring recent works to simulate, generate, or translate events.
For event simulators, we bring v2e~\cite{hu2021v2e} and DVS-Voltmeter~\cite{lin2022dvs} to simulate events. We first follow the official repository of v2e\footnote{\url{https://github.com/SensorsINI/v2e/tree/master}} to simulate events from day images with default setting. v2e also mentioned that manipulating the cutoff frequency can lead to the style change to night; following the paper, we changed the cutoff frequency to 10Hz from the default setting to simulate night events from day images. We denote this as ${\mathrm{v2e}}^{\ast}$. For more comparison, we simulate night events from night events with the default setting (${\mathrm{v2e}}^{\dagger}$). Similarly to v2e, we simulate events using DVS-voltmeter with the same process. However, we can not find a direct simulation from day to night as in v2e, so we only perform day-to-day (DVS-Voltmeter) and night-to-night (DVS-${\mathrm{Voltmeter}}^{\ast}$) simulations. We divide the simulated raw event streams into multi-bin event histograms for a fair evaluation.

\begin{table}[t]
    \centering
    \caption{Backbone of generator of each method. Several modifications are applied to networks, \eg, to enable event translation or apply conditional embedding.}
    \setlength\tabcolsep{5.2pt}
    \resizebox{0.96\linewidth}{!}{
    \begin{tabular}{l|cc|ccccc}
        \hline
        \multirow{2}{*}{Method} & \multicolumn{2}{c|}{Frame-based Generation} & \multicolumn{5}{c}{Event-to-event Translation} \\ \cline{2-8} 
         & \multicolumn{1}{c|}{EventGAN} & CMDA & \multicolumn{1}{c|}{CycleGAN} & \multicolumn{1}{c|}{GcGAN} & \multicolumn{1}{c|}{CUT} & \multicolumn{1}{c|}{UNSB} & Ours \\ \hline
        Backbone & \multicolumn{1}{c|}{U-Net} & U-Net & \multicolumn{1}{c|}{ResNet} & \multicolumn{1}{c|}{ResNet} & \multicolumn{1}{c|}{ResNet} & \multicolumn{1}{c|}{ResNet} & U-Net \\ \hline
    \end{tabular}
    }
    \label{tab:backbone}
\end{table}

EventGAN~\cite{zhu2021eventgan} needs sequential images to generate events. In other words, the distribution of events is determined by the images. Since this method entails training with ground truth, it allows the generation of day events only from day images and night events exclusively from night images. To ensure an equitable comparison, we train both models and subsequently carry out evaluations. We denote the model trained with day events as EventGAN and with night events as  ${\mathrm{EventGAN}}^{\ast}$. In the case of CMDA~\cite{xia2023cmda}, it constitutes an entire framework for day-to-night domain adaptation specialized in semantic segmentation. Still, we specifically adopted and compared only the night event generation part internally.

CycleGAN~\cite{zhu2017unpaired}, GcGAN~\cite{fu2019geometry}, CUT~\cite{park2020contrastive}, and UNSB~\cite{kim2023unpaired} employ for event-based translation, which are basically unpaired image-to-image translation methods. Some adjustments are made to translate event data. The backbone used for a generator in the image-based generation methods and the event translation models is described in~\cref{tab:backbone}.

\subsection{Applying Wavalet Decomposition to the event modality}

\begin{figure}[t!]
    \centering
    \includegraphics[width=0.99\linewidth]{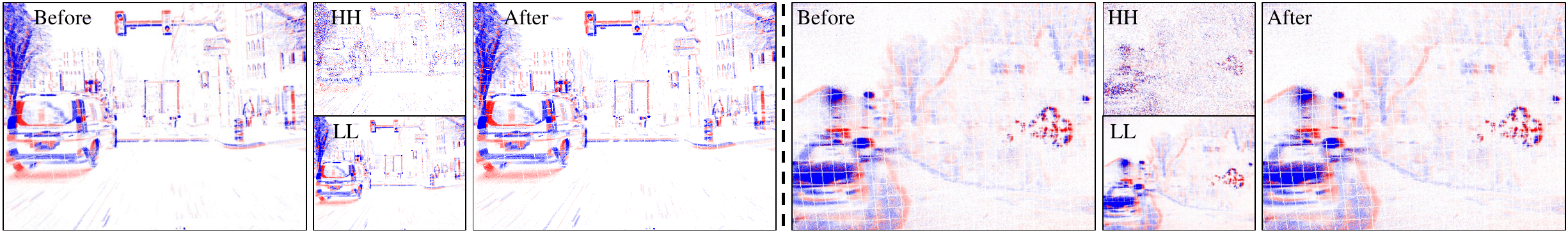}
    \caption{Applying wavelet decomposition and composition to event histograms. HH and LL denote the high- and low-frequency domains, respectively. Like images, event histograms can be divided into frequency domains through wavelet decomposition, and the original data can be restored through composition. We are motivated by the idea that the style of event data can be better transferred through processing in the high-frequency domain. Left: Day, Right: Night.}
    \label{fig:wavelet}
\end{figure}

As RGB images are disassembled into high- and low-frequency domains through decomposition, we tried to disassemble the events into high- and low-frequency domains.
We find that decomposition can also disassemble events into high- and low-frequency domains.
The overall structure is contained within the low frequency, and noises corresponding to the style of the night are predominantly distributed in the high frequency, described in~\cref{fig:wavelet}.
To reproduce the properties of events, such as natural noise, we are motivated by the idea that the style of event data can be better transferred through processing in the high-frequency domain.

\subsection{Dataset}
\textbf{DSEC datset}~\cite{Gehrig21ral} provides data on various driving scenes from their sensor system. It comprises stereo image pairs and raw event data, documenting diverse urban and rural environments at multiple times. The dataset also includes semantic segmentation labels, object bounding boxes, and other annotations. 
By leveraging extrinsic parameters for warping, we can produce high-quality event and pixel-wise aligned images, both with a resolution of 640 $\times$ 480.

We divided the DSEC sequences into day and night with a specific classification criterion. We developed the criteria by taking into account both the images and events within a particular sequence. In the DSEC dataset, daytime is distinctly characterized by a bright sky in an image and minimal irregular noise on a road surface in event modality.
Nighttime sequences contain a dark sky of an image, significant ground noise, and evident streetlight influences in events.
Nevertheless, the mere presence of streetlights or the occurrence of ground noise does not consistently indicate nighttime conditions. Additional factors, such as the rugged texture of events, are also accounted for. Days characterized by too much cloudy skies or insufficient evidence of nighttime are categorized as `others' and are excluded from the training dataset.
The event sequences are stratified into train and test splits, further stratified into daytime and nighttime subsets. Subsequently, each sequence is precisely partitioned at intervals of 100ms to formulate the event stream. The number of event streams for each sequence, as determined through this process, is presented in~\cref{tab:seq}.

We chose the DSEC dataset as our primary dataset because it provides both day and night event streams, which are easily discernible and suitable for unpaired translation. Additionally, the DSEC dataset includes a substantial volume of day and night events accompanied by aligned images. Comparable data is required to evaluate our proposed method against recent approaches. With the DSEC dataset, we can conduct not only event-to-event translation but also event generation and simulation using images captured at the same timestamp.

\begin{table}[t]
\centering
\caption{The details about the train/test sequence splits.}
\setlength\tabcolsep{8.2pt}
\resizebox{.78\linewidth}{!}{
\begin{tabular}{l|l|lll}
\hline
\textbf{Split} & \textbf{Time} & \textbf{Sequence} & \textbf{\# event streams} & \textbf{Total} \\ \hline
\textbf{Train} & \textbf{Day} & interlaken\_00\_c & 267 & \textbf{5774} \\
 &  & interlaken\_00\_d & 994 &  \\
 &  & interlaken\_00\_e & 994 &  \\
 &  & zurich\_city\_04\_a & 349 &  \\
 &  & zurich\_city\_04\_b & 133 &  \\
 &  & zurich\_city\_04\_c & 589 &  \\
 &  & zurich\_city\_05\_a & 875 &  \\
 &  & zurich\_city\_05\_b & 813 &  \\
 &  & zurich\_city\_06\_a & 760 &  \\ \cline{2-5} 
 & \textbf{Night} & zurich\_city\_03\_a & 440 & \textbf{4598} \\
 &  & zurich\_city\_09\_a & 905 &  \\
 &  & zurich\_city\_09\_b & 182 &  \\
 &  & zurich\_city\_09\_c & 660 &  \\
 &  & zurich\_city\_09\_d & 848 &  \\
 &  & zurich\_city\_09\_e & 407 &  \\
 &  & zurich\_city\_10\_a & 1156 &  \\ \hline
\textbf{Test} & \textbf{Day} & zurich\_city\_07\_a & 730 & \textbf{1895} \\
 &  & zurich\_city\_11\_c & 977 &  \\
 &  & zurich\_city\_13\_a & 188 &  \\ \cline{2-5} 
 & \textbf{Night} & zurich\_city\_02\_c & 1440 & \textbf{3022} \\
 &  & zurich\_city\_10\_b & 1201 &  \\
 &  & zurich\_city\_12\_a & 381 &  \\ \hline
\end{tabular}}
\label{tab:seq}
\end{table}

Furthermore, alternative datasets (\eg, N-Caltech101~\cite{ncaltech}) incorporate synthetic components obtained from event cameras capturing monitor panels rather than authentic 3D environments. Consequently, we opt for the DSEC dataset, which offers images and events depicting real-world day-night scenarios. Moreover, leveraging the comprehensive research conducted on event camera applications within the DSEC dataset, which also supplies labels, enables us to conduct segmentation and detection tasks that necessitate pertinent annotations.

\noindent
\textbf{MVSEC dataset.}
Additionally, we employ the MVSEC dataset~\cite{zhu2018multivehicle} to evaluate the model's ability to generalize to unseen datasets. Similar to the DSEC dataset, MVSEC captures driving scenes featuring day and night events with aligned images. However, notable differences exist in the data distributions between them due to varying environments and camera sensors. Specifically, the MVSEC dataset is derived from diverse event cameras with distinct resolutions, sensor sizes, event trigger thresholds, and deployment locations. These variations make it suitable for assessing the model's ability to generalize to unseen scenarios using the MVSEC dataset.

We use `outdoor\_day1' sequence as day events to perform the unseen translation. The `outdoor\_night1' sequence is used as a reference group.

\section{Evaluation Metric}
\label{sec:metric}

\subsection{Details about the Evaluation Metric}
As mentioned in the main paper, for the direct evaluation of generated events, we utilize the Fréchet Inception Distance (FID) score~\cite{heusel2017gans} and the Fréchet Video Distance (FVD)~\cite{unterthiner2018towards} score, commonly used in the image society.

They calculate feature distance using pre-trained models which are trained on the ImageNet dataset~\cite{deng2009imagenet}.
However, these existing pre-trained models were initially trained on image modality data, therefore, they cannot be directly applied to event data. To utilize these metrics to evaluate events, we trained the Inception-v3~\cite{szegedy2016rethinking} and 3D Convnet (I3D)~\cite{carreira2017quo} on the event modality. We trained models with DSEC day-night classification (two classes), left behind N-ImageNet~\cite{kim2021n}, an event modality version of ImageNet data. Analysis related to this will be covered in the following sub-section.

\begin{figure}[th!]
    \centering
    \includegraphics[width=0.85\linewidth]{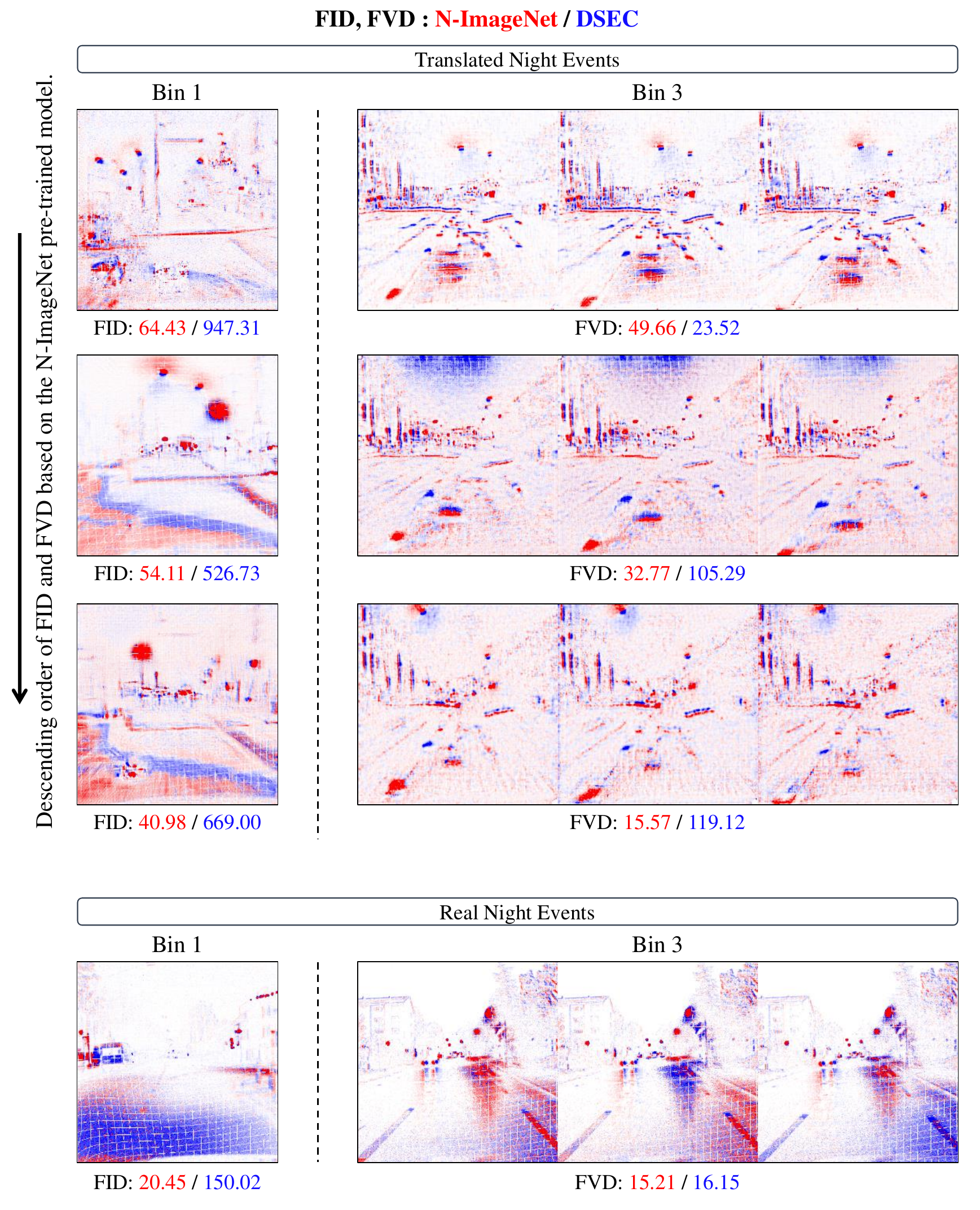}
    \caption{Comparisons of the FID and FVD scores using the pre-trained networks on N-ImageNet~\cite{kim2021n} and DSEC~\cite{Gehrig21ral} datasets. \textcolor{red}{The values on the left are obtained from a pre-trained model on N-ImageNet}, while \textcolor{blue}{the values on the right are obtained from a model trained on DSEC}, respectively, indicated as N-ImageNet/DSEC. The metrics are sorted in descending order based on FID and FVD by N-ImageNet pre-trained model. The bottom row represents a real night event.}
    \label{fig:metric_comparison}
\end{figure}

Although FID and FVD can effectively distinguish the feature distribution between day and night, it is essential to assess whether the generated events retain the spatio-temporal characteristics of real events and can be reliably utilized in downstream tasks. Hence, we evaluate the generated events by converting them into image format using a pre-trained E2I network. The E2I conversion is performed by the E2VID network~\cite{rebecq2019high}, trained on data comprising night scenes.
To train the E2I network, we split the night sequences into two subsets: a training set and a validation set. For the training set, we utilized four sequences:  `zurich\_city\_03\_a', `zurich\_city\_09\_a', `zurich\_city\_09\_c', and `zurich\_city\_09\_e.' For the validation set, we selected `zurich\_city\_09\_b', `zurich\_city\_09\_d', and `zurich\_city\_10\_a'. We carefully selected these sequences to ensure that there is no overlap between the training set used for E2I and the validation set used in the translation task. To perform E2I for each bin, we trained a separate E2I model for each corresponding bin.

For the E2I evaluation, we utilize both FID and Kernel Inception Distance (KID) scores~\cite{bińkowski2018demystifying}. The KID metric quantifies the similarity between real and generated samples by computing the squared Maximum Mean Discrepancy with a polynomial kernel. We denote the two metrics for E2I evaluation as E2I-FID and E2I-KID. All values of the E2I-KID score in the main paper and supplementary material are multiplied by 100.

To assess the outputs, we establish distinct reference groups for each bin. We compare FID and FVD against real night events. In the case of E2I-FID/KID, we utilize the results obtained from translating night events into images using the E2I network as the reference group for each bin. For example, to evaluate our results using E2I-FID for bin 8, one side of the evaluation network integrates the E2I transformation of our outputs, while the other side incorporates the results of translating night events from the test set into images. The evaluation is then conducted accordingly.

\subsection{Analysis of FID and FVD metrics based on the training dataset}
The rationale behind opting for day-night classification within DSEC instead of object classification within N-ImageNet~\cite{kim2021n} as a feature metric is underpinned by the following factors: (1) In the training regimen utilizing N-ImageNet, the neural network is never exposed to natural night event streams, particularly those encompassing varied illuminance fluctuations and noise occurrences, such as driving scenarios. Consequently, when trained on N-ImageNet, the night event stream represents unobserved data, rendering the accurate discrimination of the day-night distribution as a feature metric insufficient. (2) Our designated task, distinct from a generative task, pertains specifically to day-night translation, entailing characteristics that are not easily acquired solely through conventional object classification. While a fixed distribution of RGB values may suffice in image modality, events necessitate consideration of spatio-temporal features associated with motion. Hence, tailored day-night feature metric-based learning is indispensable for effective discrimination.

Figure~\ref{fig:metric_comparison} provides empirical support for the aforementioned claims. Our assessments involved the evaluation of translated night events and authentic night events using networks pre-trained on N-ImageNet with 1000 classes. Upon examining the bottom row, it becomes apparent that when trained on N-ImageNet, both FID and FVD scores for real night events register low values, indicating accurate assessments for True/Positive instances. According to the intended design, an improvement in quality should be discernible when assessing translated events further down the row based on N-ImageNet-derived metrics. However, irrespective of actual quality, artificially introducing excessive noise into the background can result in favorable metric outcomes. We attribute this phenomenon to the fact that the pre-trained feature-metric network has not been exposed to such dynamic nocturnal data during the training phase. In contrast, the network trained on DSEC for day-night classification consistently generates favorable metric values for genuine night events, and instances with excessive background noise fail to yield favorable FID and FVD results.

Given these observations, we decided against directly utilizing conventional metrics from the image domain, such as training on a large-scale object classification dataset (\eg, ImageNet~\cite{deng2009imagenet}), due to its task-agnostic nature concerning the target task. Instead, we adapted our training approach to align with unpaired day-night event translation and selected metrics accordingly.

\subsection{Subjective Tests for the Evaluation Metric}
\begin{figure}[t]
    \centering
    \includegraphics[width=0.99\linewidth]{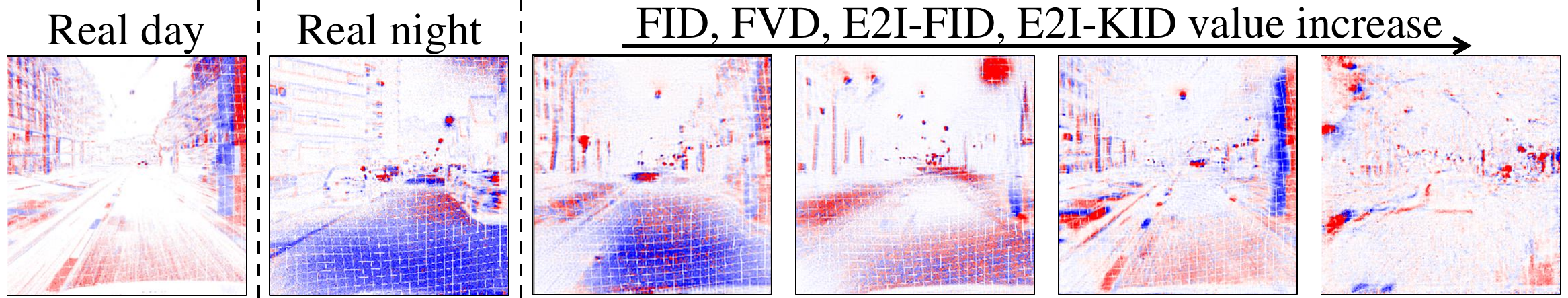}
    \caption{A sample of subjective tests. They were shown references, real events, and then asked to rank four translated samples by closeness to night events.}
    \label{fig:subjective_test}
\end{figure}

\begin{table}[t]
    \centering
    \caption{Quantitative result of subjective tests. We compare the results with top 1 accuracy, total accuracy, and cosine similarity.}
    \setlength\tabcolsep{10.2pt}
    \resizebox{0.75\linewidth}{!}{
    \begin{tabular}{c|c|c|c|c}
    \hline
     & FID & FVD & E2I-FID & E2I-KID \\ \hline
    Top1 Acc. (\%) & 85.0 & 85.0 & 80.0 & 90.0 \\ \hline
    Acc. (\%) & 75.6 & 73.8 & 76.3 & 71.3 \\ \hline
    Cos. Sim. (\%) & 95.9 & 84.9 & 89.2 & 80.7 \\ \hline
    \end{tabular}}
    \label{tab:subjective_test}
\end{table}

We recruited 20 volunteers to validate our metrics. As shown in~\cref{fig:subjective_test}, we provided testers with real-day events and the results of translating these events using different methods, as well as reference night events. We then asked them to rank which created events were most similar to the reference night events. We compared their rankings with those from our metrics in~\cref{tab:subjective_test}. Our metrics closely align with human perception.

\section{More Analysis and Discussion}
\label{sec:discussion}
\subsection{Model Parameters}
We provide the model parameters for various bin sizes in Table~\ref{tab:parm_gpu}. 
As mentioned in Sec.~\ref{sec:imple_details}, to effectively handle translations for larger bin sizes, the channel dimension must increase, resulting in a higher number of parameters. 
Our method, compared to previous GAN approaches, has a small number of parameters. Even with the addition of various proposed modules over the diffusion GAN, the portion of the total number of parameters remains small, confirming that our methodologies have been efficiently designed.

\begin{table}[t]
    \centering
    \caption{Model parameters.}
    \setlength\tabcolsep{6.2pt}
    \resizebox{.6\linewidth}{!}{
    \begin{tabular}{cccc}
    \hline
    Params (M) & CycleGAN & UNSB & Ours \\ \hline
    2 Channel (Bin = 1) & 28.272 & 21.528 & 24.748 \\
    6 Channel (Bin = 3) & 34.372 & 25.502 & 28.273 \\
    16 Channel (Bin = 8) & 56.966 & 39.874 & 44.207 \\ \hline
    \end{tabular}
    \label{tab:parm_gpu}}
\end{table}

\subsection{Hyperparameter Analysis}
\label{sec:temperature}
We conducted an additional experiment to determine the appropriate value for the temperature of temporally shuffling contrastive regularization, denoted as $\tau$ (Eq. (\textcolor{red}{9}) in the main paper). The results are summarized in Table~\ref{tab:ablation_temperature}. In this experiment, we maintained the model's configuration constant while varying only the temperature parameter. We explored temperature values ranging from 0.01 to 0.15 at intervals of 0.02.
Across various temperature variations, our method consistently outperformed other techniques in terms of E2I-FID and E2I-KID scores.
However, when considering the FVD score, our approach showcased enhanced performance only at specific temperatures, notably 0.11 and 0.13.
These temperatures also yielded more favorable results across all metrics. Since the FVD metric evaluates the quality of generated events concerning nighttime scenes, we infer that the temperature of temporal contrastive learning influences the quality of events by regulating their temporal continuity.
Consequently, we opted for a final temperature of 0.11, which consistently yielded superior results across the evaluated metrics compared to the other temperature options.

\begin{table}[h]
    \centering
    \caption{Quantitative result of different temperature of temporally shuffling contrastive regularization. The temperature of 0.11 showed significantly better performance compared to other options.}
    \setlength\tabcolsep{7.2pt}
    \resizebox{.55\linewidth}{!}{%
    \begin{tabular}{c|ccc}
        \hline
        Temperature  & FVD$\downarrow$ & E2I-FID$\downarrow$ & E2I-KID$\downarrow$ \\ \hline
        0.01 &   80.12 & \underline{20.45} & \underline{1.59} \\
        0.03 &   77.33 & 21.56 & 1.60 \\
        0.05 &   84.33 & 22.34 & 4.62 \\
        0.07 &   60.43 & 24.70 & 3.07 \\
        0.09 &   91.98 & 22.46 & 1.97 \\
        0.11 &   \underline{7.19} & \textbf{18.01} & \textbf{1.24} \\
        0.13 &   \textbf{5.67} & 22.46 & 1.91 \\
        0.15 &   51.87 & 22.92 & 2.64 \\ \hline
    \end{tabular}}
    \label{tab:ablation_temperature}
\end{table}

\section{More about Qualitative Results}
\label{sec:more_qual}
In DSEC, the ground-truth for downstream tasks is aligned with rectified images. Consequently, we rectified raw events during training, leading to grid effects. 
As seen in figures of the main paper, grid artifacts are also observed in other translation methodologies and are not a specific characteristic of our approach.
In the above figure, training on DSEC using events with and without rectification affects the output based on the rectification of input events.
Afterward, inference on MVSEC shows no grid artifacts, indicating that a proper training strategy can resolve this issue.

\begin{figure}[t]
    \centering
    \includegraphics[width=0.99\linewidth]{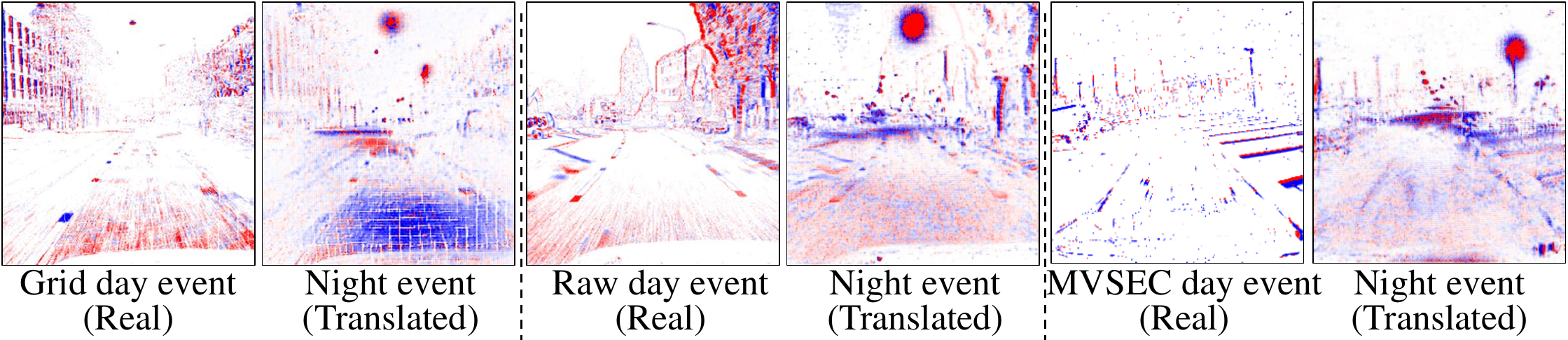}
    \caption{Grid Effect. We train the model using rectified and raw events together. As the rectified events are used for input of the model during the inference, results also contain the grid. On the other hand, as the raw events are used for input of the model during the inference, no grid are represented. }
    \label{fig:grid}
\end{figure}

\begin{figure}[p]
    \centering
    \includegraphics[width=.99\linewidth]{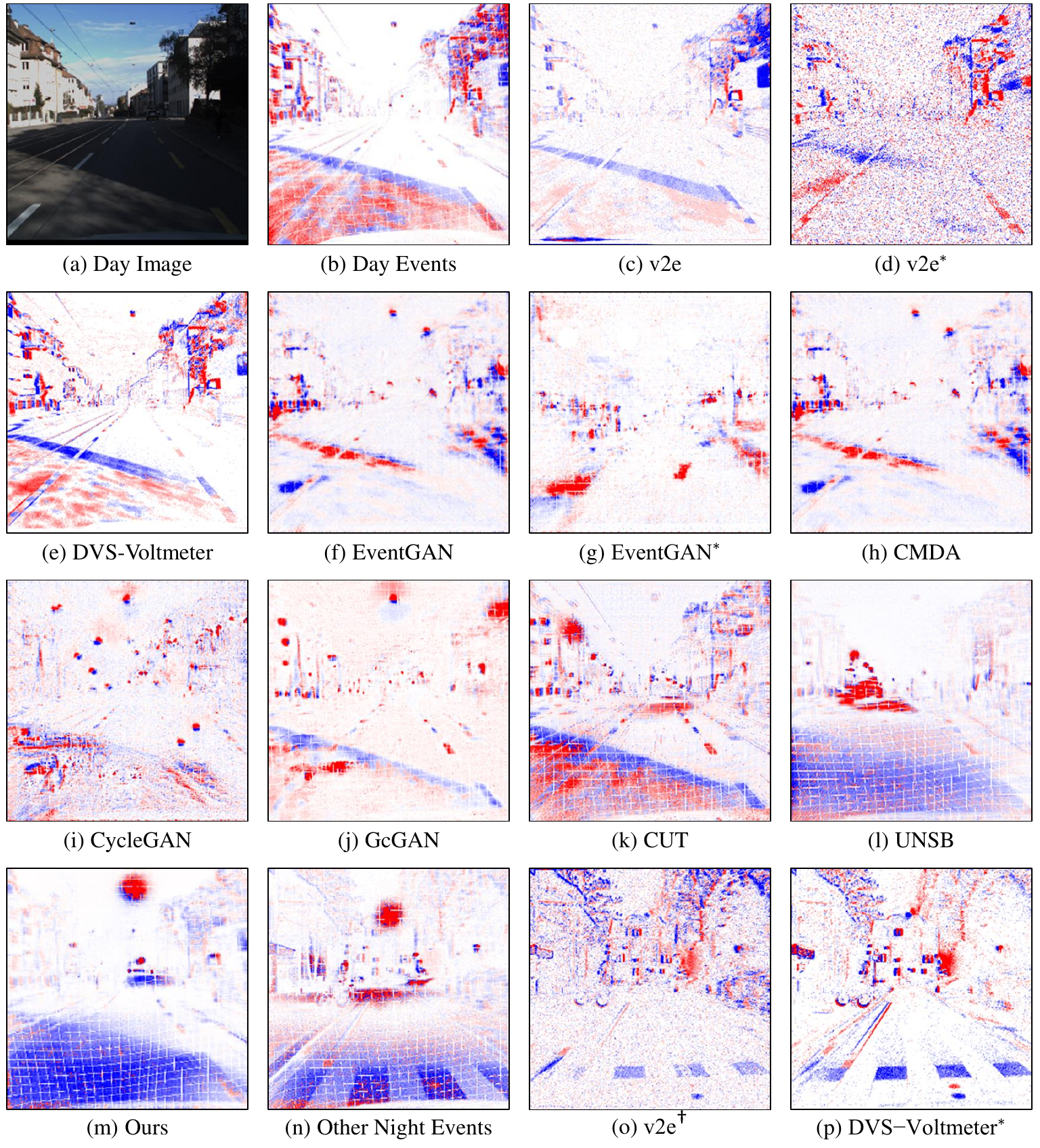}
    \caption{Qualitative comparison for bin 1. The day images and aligned day events are placed in (a) and (b), respectively. (c) to (m) is simulated/generated/translated night events from day events. (n) is other night events as a reference. For (o) and (p), ${\mathrm{v2e}}^{\dagger}$ and DVS-${\mathrm{Voltmeter}}^{\ast}$ simulate with night events, so we separate them from others. These two results are simulated from (n). The result of our proposed method produces the night events style with clear background as the reference night events have.}
    \label{fig:bin1_1}
\end{figure}

\begin{figure}[p]
    \centering
    \includegraphics[width=.99\linewidth]{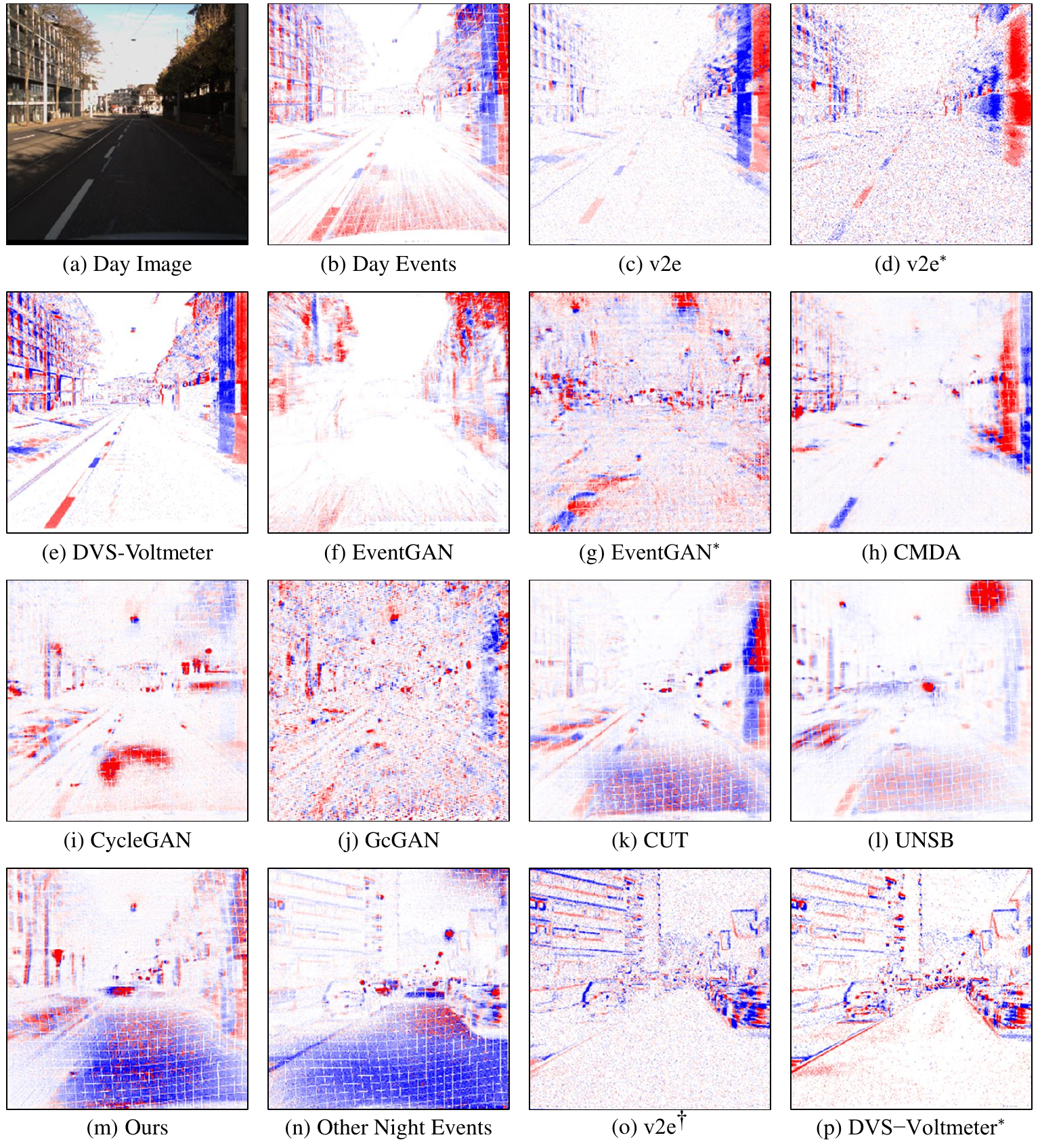}
    \caption{Qualitative comparison for bin 1. The day images and aligned day events are places in (a) and (b), respectively. (c) to (m) is simulated/generated/translated night events from day events. (n) is other night events as a reference. For (o) and (p), ${\mathrm{v2e}}^{\dagger}$ and DVS-${\mathrm{Voltmeter}}^{\ast}$ simulate with night events, so we separate them from others. These two results are simulated from (n). The result of our proposed method produces the night events style with clear background as the reference night events have.}
    \label{fig:bin1_2}
\end{figure}

\begin{figure}[p]
    \centering
    \includegraphics[width=1\linewidth]{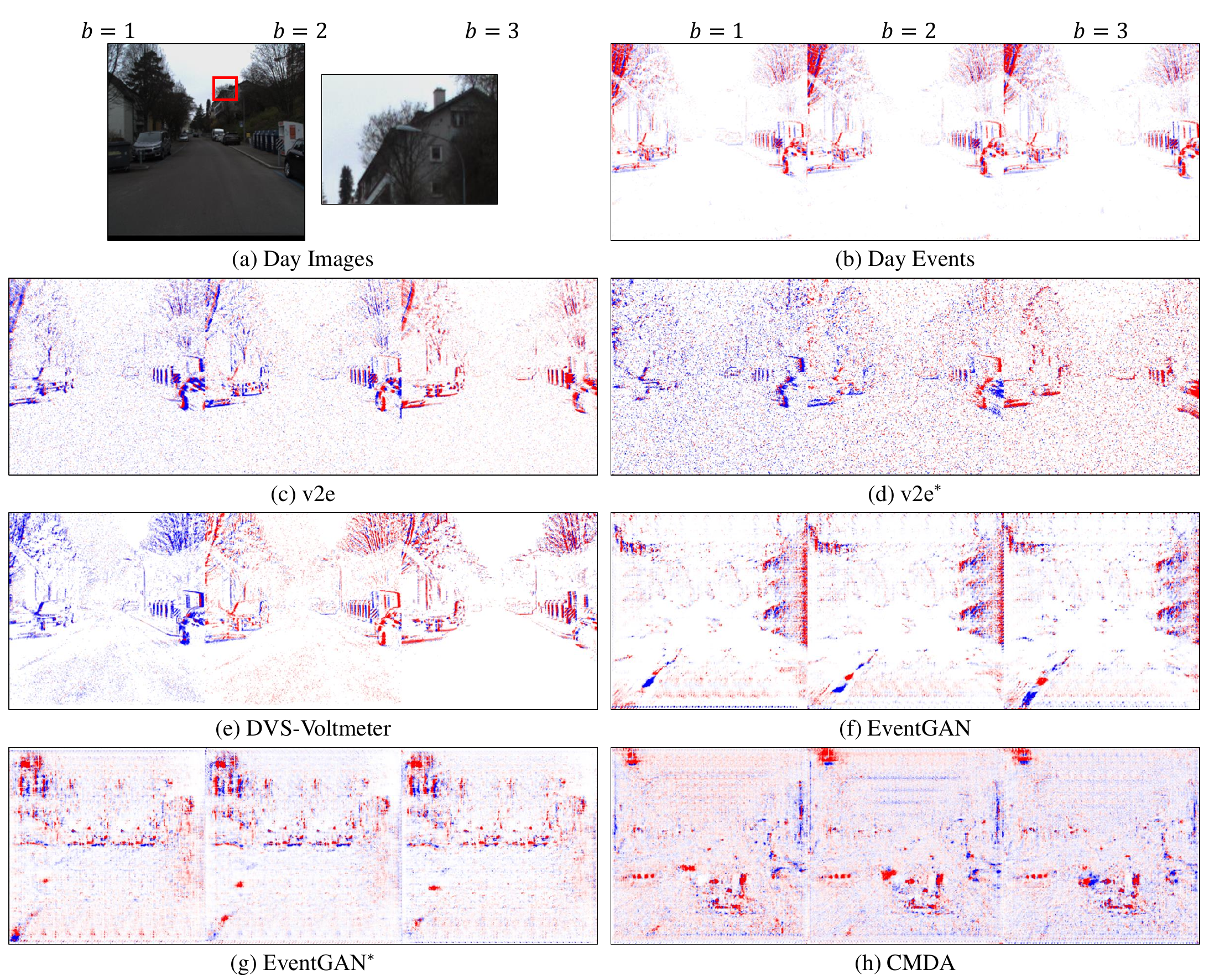}
    \caption{Qualitative comparison for bin 3. Please refer to~\cref{fig:bin3_2_2} for the ongoing visualization.}
    \label{fig:bin3_2_1}
\end{figure}

\begin{figure}[p]
    \centering
    \includegraphics[width=1\linewidth]{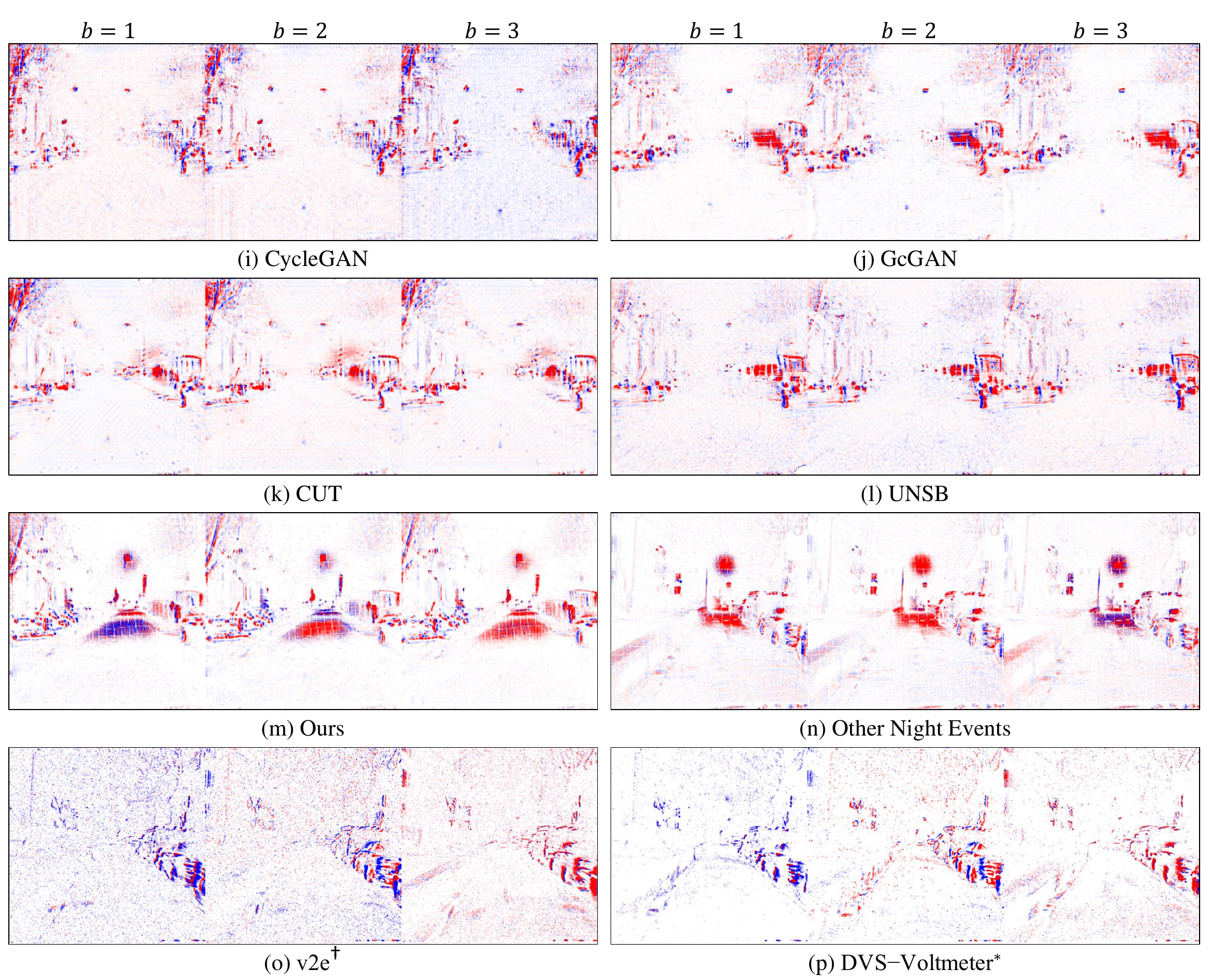}
    \caption{Qualitative comparison for bin 3. (a) to (h) is placed in~\cref{fig:bin3_2_1}. 
    The day images and aligned day events are represented by (a) and (b), respectively. Simulated night events from day events are shown in (c) to (e). The visualizations of image-based generation methods are illustrated by (h) and (g), while (i) to (m) depict the results of each event-based translation method. Reference night events are denoted by (n). In (o) and (p), ${\mathrm{v2e}}^{\dagger}$ and DVS-${\mathrm{Voltmeter}}^{\ast}$ simulate night events, separated from others and derived from (n). The result of our proposed method produces night events with a clear background, similar to the reference night events.}
    \label{fig:bin3_2_2}
\end{figure}

\begin{figure}[p]
    \centering
    \includegraphics[width=1\linewidth]{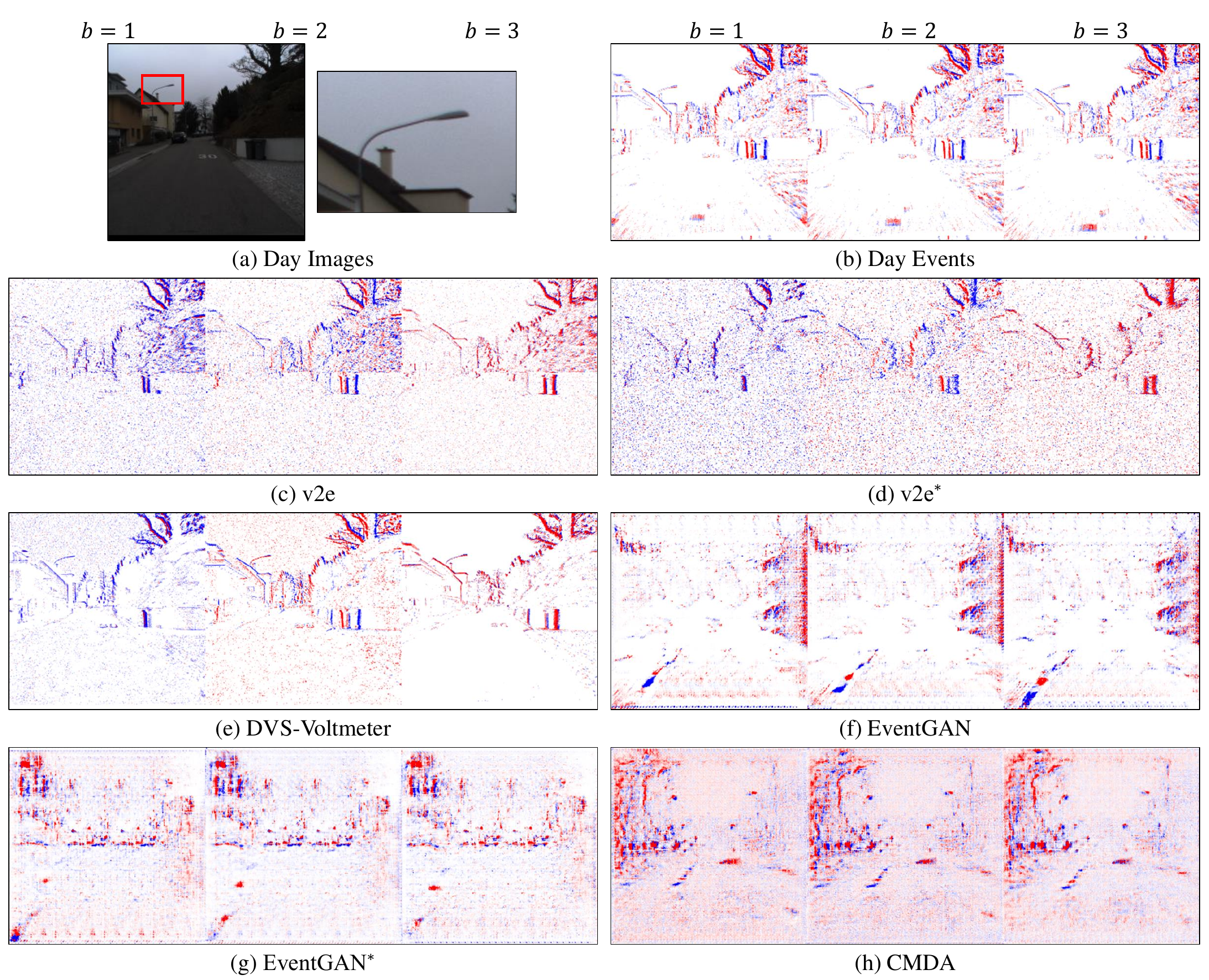}
    \caption{Qualitative comparison for bin 3. Please refer to~\cref{fig:bin3_1_2} for the ongoing visualization.}
    \label{fig:bin3_1_1}
\end{figure}

\begin{figure}[p]
    \centering
    \includegraphics[width=1\linewidth]{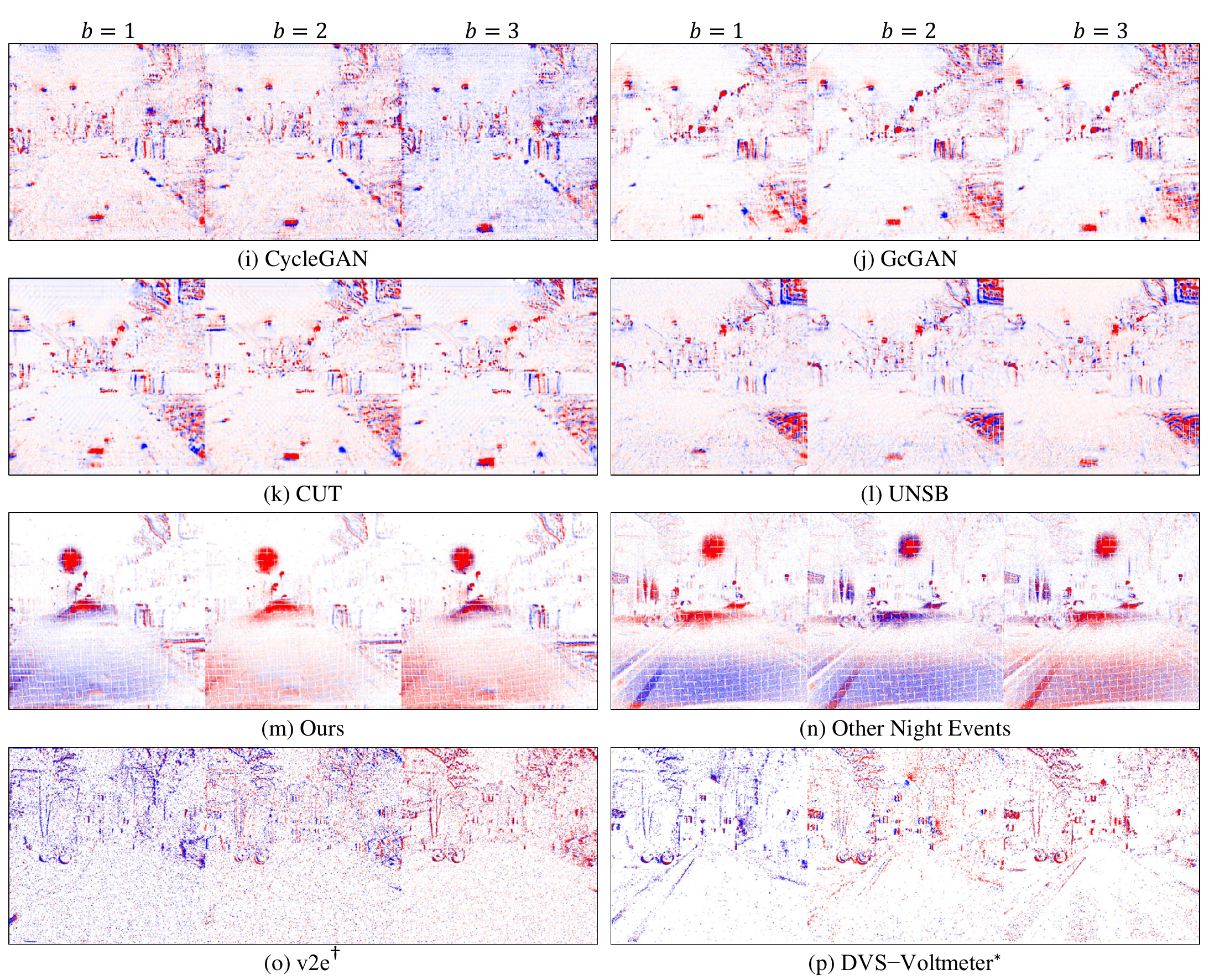}
    \caption{Qualitative comparison for bin 3. (a) to (h) is placed in~\cref{fig:bin3_1_1}. The day images and aligned day events are represented by (a) and (b), respectively. Simulated night events from day events are shown in (c) to (e). The visualizations of image-based generation methods are illustrated by (h) and (g), while (i) to (m) depict the results of each event-based translation method. Reference night events are denoted by (n). In (o) and (p), ${\mathrm{v2e}}^{\dagger}$ and DVS-${\mathrm{Voltmeter}}^{\ast}$ simulate night events, separated from others and derived from (n). The result of our proposed method produces night events with a clear background, similar to the reference night events.}
    \label{fig:bin3_1_2}
\end{figure}

\begin{figure}[p]
    \centering
    \includegraphics[width=1\linewidth]{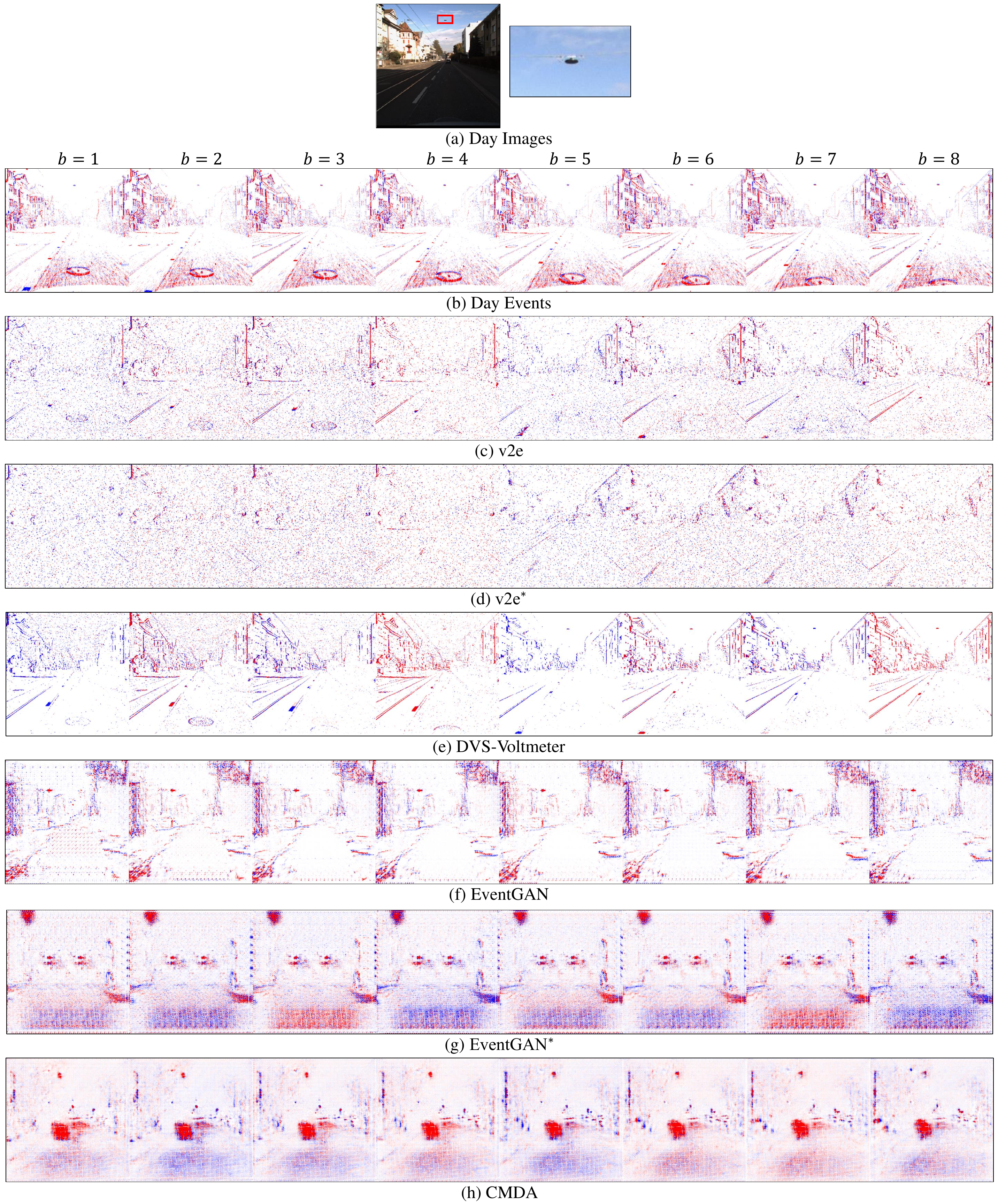}
    \caption{Qualitative comparison for bin 8. Please refer to~\cref{fig:bin8_1_2} for the ongoing visualization.}
    \label{fig:bin8_1_1}
\end{figure}

\begin{figure*}[p]
    \centering
    \includegraphics[width=1\linewidth]{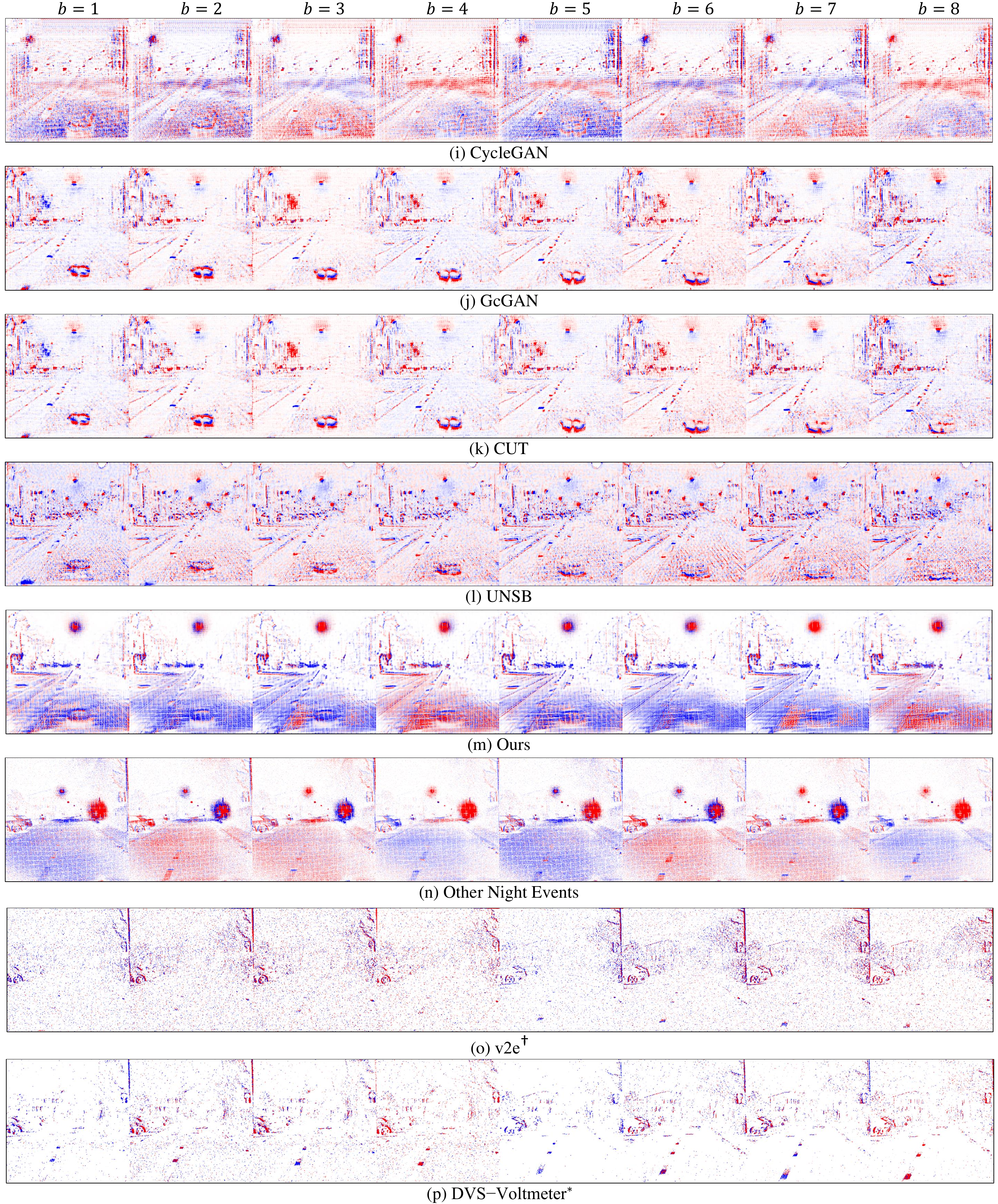}
    \caption{Qualitative comparison for bin 8. (a) to (h) is placed in~\cref{fig:bin8_1_1}. 
    The day images and aligned day events are represented by (a) and (b), respectively. Simulated night events from day events are shown in (c) to (e). The visualizations of image-based generation methods are illustrated by (h) and (g), while (i) to (m) depict the results of each event-based translation method. Reference night events are denoted by (n). In (o) and (p), ${\mathrm{v2e}}^{\dagger}$ and DVS-${\mathrm{Voltmeter}}^{\ast}$ simulate night events, separated from others and derived from (n). The result of our proposed method produces night events with a clear background, similar to the reference night events.}
    \label{fig:bin8_1_2}
\end{figure*}

We provide additional qualitative comparisons. We add all visualized results of every method. For bin 1, please refer to~\cref{fig:bin1_1} and \cref{fig:bin1_2}. Also, for bin 3, unlike the main paper, which showed specific bins, we present results for all bins, allowing for the examination of temporal flow. \Cref{fig:bin3_1_1}, \cref{fig:bin3_1_2}, \cref{fig:bin3_2_1}, and \cref{fig:bin3_2_2} represent the result of bin 3. 
\Cref{fig:bin3_1_1} and \cref{fig:bin3_1_2}, as well as \cref{fig:bin3_2_2} and \cref{fig:bin3_2_1}, each represent results for the same scene, respectively. We also show the result of bin 8 in~\cref{fig:bin8_1_1} and \cref{fig:bin8_1_2}. 
 
Clearly, based on the qualitative results, other methods fail to generate realistic night events from two main perspectives: (1) severe distortion of structure or an excessive attempt to maintain structure, resulting in a lack of distinct night characteristics, and (2) failure to recognize and learn the separated characteristics between day and night events, leading to excessive generation of events that do not fit appropriately into the background of the generated results.
In contrast, our approach effectively preserves the spatio-temporal characteristics of the night events while seamlessly adapting the structural features from day to night. Instead of simply generating excessive noise, our method generates realistic night event noise with leaving unnecessary backgrounds empty and maintaining the structure. These differences become apparent during event translations around artificial lighting sources, such as streetlights.

\section{Details of Applications}
\label{sec:application}
As we mentioned in the main paper, to check the effectiveness of translation, we train several downstream networks with day and translated night events with annotations. As a control group, we train the network only with day events and evaluate with real night events. For others, we translate several annotated day sequences to become night events. As a train set, we select the following sequences: `interlaken\_00\_c', `interlaken\_00\_e', `zurich\_city\_04\_a', `zurich\_city\_04\_b', `zurich\_city\_04\_c', `zurich\_city\_05\_a', `zurich\_city\_0
5\_b', `zurich\_city\_06\_a', `zurich\_city\_07\_a', `zurich\_city\_11\_c', `zurich\_ci
ty\_13\_a'.
Excluding a training scenario involving only day events, the events from the respective sequences, `zurich\_city\_07\_a', `zurich\_city\_11\_c', and `zurich\_city\_13\_a', are replaced with events translated by each method during the training process.

We incorporate the E2VID network\cite{rebecq2019high} into our downstream task. This evaluation was limited to bin 3, and the results are presented in Table~\textcolor{red}{5} of the main paper. To evaluate the outcomes, we utilize SSIM~\cite{wang2004image}. SSIM measures the structural similarity between two different images. Given our aim of generating images with stylistic variations from the ground truth image, we deem indirect metrics like SSIM more appropriate than direct comparisons.

\subsection{Nighttime Event-based Semantic Segmentation.}
\begin{table}[h]
    \centering
    \caption{Label mapping for Nighttime Event-based Semantic Segmentation.}
    \setlength\tabcolsep{5.2pt}
    \resizebox{.99\linewidth}{!}{%
    \begin{tabular}{c|ccccccccccc}
    \hline
    Class  &  background & building & fence &
    person & pole & road & sidewalk & vegetation & car & wall & traffic sign
    \\ \hline
    ID & 0 &  1 & 1& 2& 3& 4 & 5 & 6 & 2 & 1 & 3
    \\ \hline
    \end{tabular}}
    \label{tab:class_category_semseg}
\end{table}
\noindent
Gathering annotated event data for nighttime is challenging as existing labeled datasets mostly contain daytime data. Models trained on daytime datasets may underperform in the night domain. One solution is to complement nighttime data with annotated day events through day-to-night translation, thereby fully exploiting the available annotations. Then, we translate labeled day events into night events while preserving the annotated segmentation mask.

\begin{figure}[p]
    \centering
    \includegraphics[width=0.99\linewidth]{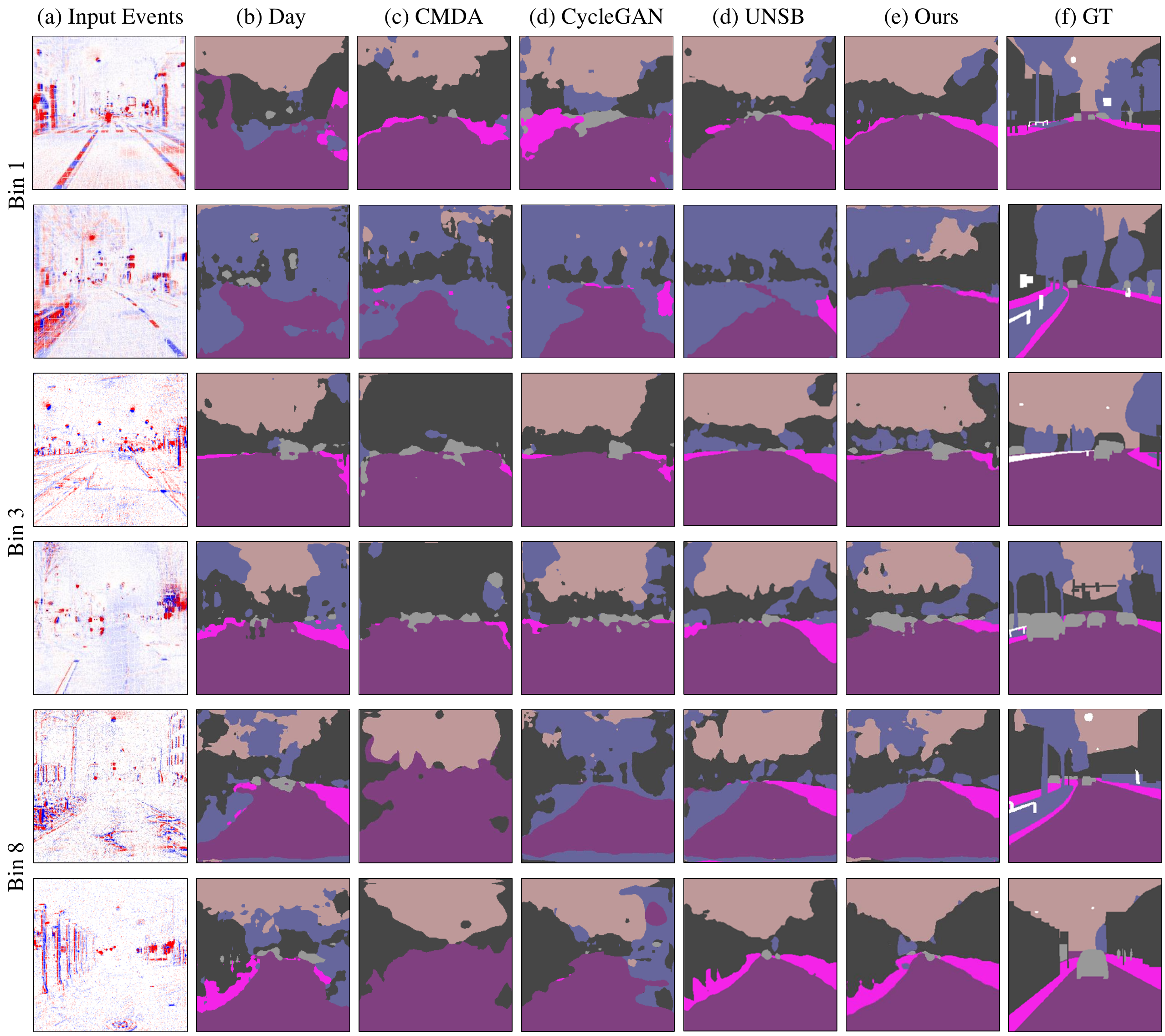}
    \caption{Visual comparisons of semantic segmentation results using real night events. ``Day'' signifies training solely with day events as a reference group. Other approaches leverage both day and translated night events during training which are annotated. Notably, our method demonstrates robustness in handling noise inherent in night events, leading to enhanced object boundary delineation and reduced artifact presence.}
    \label{fig:seg_result}
\end{figure}

To train the segmentation network, we employ DeepLabv3+~\cite{chen2018encoder} as the decoder and ResNet-34~\cite{he2016deep} as the backbone in the semantic segmentation model. 
For training, we utilize the labels provided by ESS~\cite{sun2022ess} for daytime events, and for testing, we use the labels provided by CMDA~\cite{xia2023cmda} for nighttime events. 
In training, we train the segmentation network using only the labels of ESS overlapping with the day of the test sequence in~\cref{tab:seq}. In addition, to fully leverage the benefits of domain translation, we train the network using both day events and the generated night events. 
Some classes are absent in training and test sequences, and others have insufficient instances, potentially introducing bias. To address this, we combine the 11 classes provided by ESS into 7 classes by merging them with the high-level classes provided by Cityscapes~\cite{cordts2016cityscapes} and conduct evaluations based on these 7 classes. We present the newly mapped labels in Table~\ref{tab:class_category_semseg}. The qualitative results are presented in Tab. \textcolor{red}{6} of the main paper. For more analysis, only the model utilizing events translated through our proposed method perform better than the model solely trained by day events. As the bin size increases, temporal rules become crucial, and while other methods struggle to handle this, our approach, with the temporally shuffling contrastive regularization and temporally disentangled encoder, handles the properties and preserves the continuity well. In Fig.~\ref{fig:seg_result}, we provide more qualitative results of nighttime event-based semantic segmentation using day-to-night translation.

\subsection{Nighttime Event-based Obejct Detection.}
Events captured in nighttime driving conditions are challenging to accurately delineate structures' boundaries due to high noise levels and low lighting conditions. It is challenging to draw object boundary boxes in nighttime events. Therefore, in object detection networks, typically, training on nighttime events is either a small portion of the entire dataset or nearly absent. Subsequently, we employ translation on annotated day events to create night events, ensuring the preservation of object detection annotations. To train the object detection network, we adopt Feature Pyramid Network~\cite{lin2017feature} with ResNet-101~\cite{he2016deep} as the backbone. We only utilize events as input; there is no use of images.
For training, we employ the annotations provided by \cite{zhou2023rgb} with 3 classes: person, large vehicle, and car. We conducted experiments on our method, EventGAN~\cite{zhu2021eventgan}, CMDA~\cite{xia2023cmda}, CycleGAN~\cite{zhu2017unpaired}, and UNSB~\cite{kim2023unpaired}. The control group trained the detection model using only daytime events for comparison. 

\begin{figure}[p]
    \centering
    \includegraphics[width=0.99\linewidth]{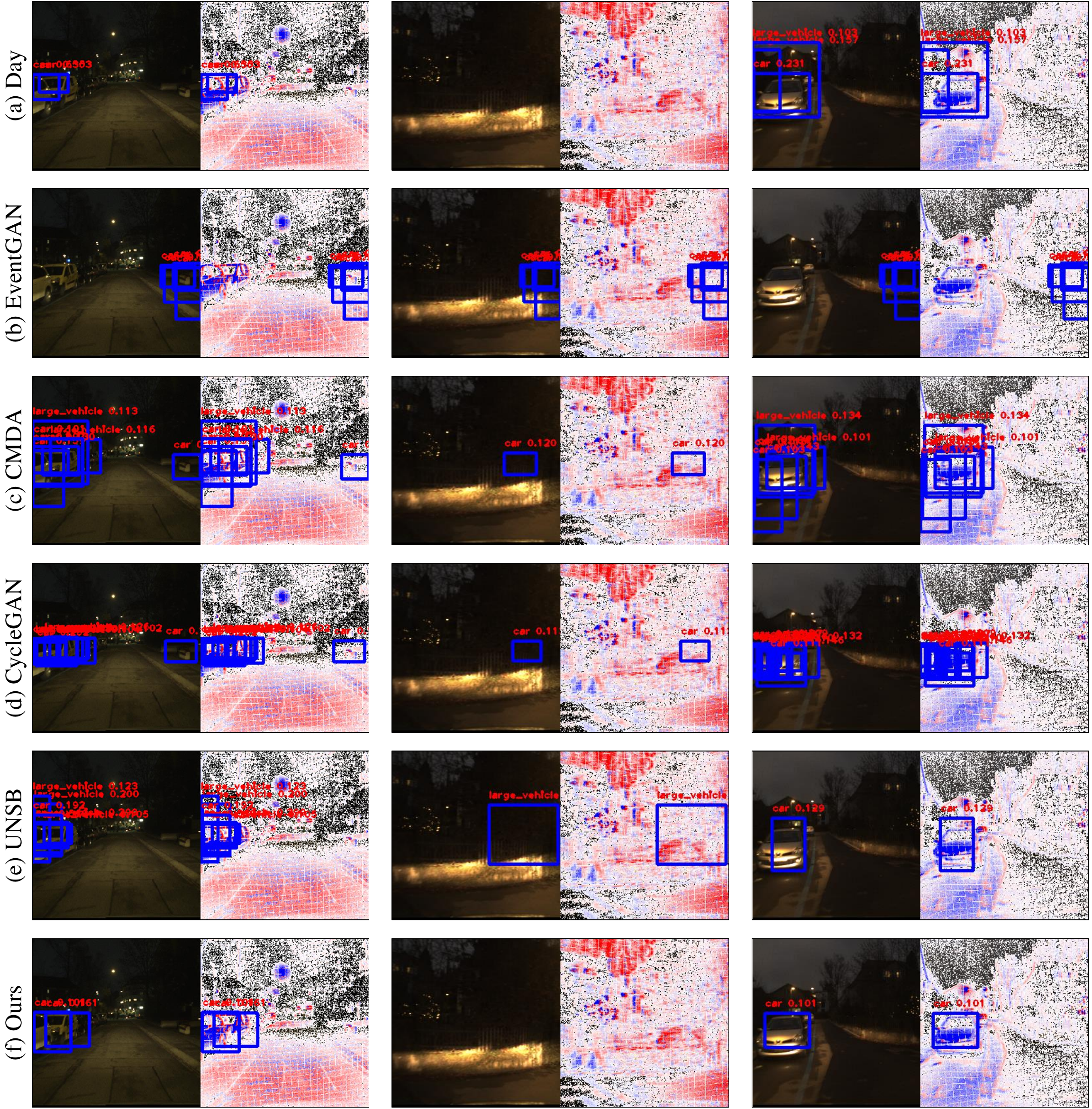}
    \caption{Visual comparisons of object detection results using real night events. We visualize aligned image and events together with bounday boxes estimation results. ``Day'' signifies training solely with day events as a reference group. Other approaches leverage both day and translated night events during training which are annotated. Our method is robustly trainable to the noise in night events, enabling better identification of object boundaries with fewer artifacts. While other methods may fail to locate an object or search in the wrong locations correctly, our approach accurately detects the presence of an object and refrains from detection when it is absent.}
    \label{fig:ob_result}
\end{figure}

We perform object detection evaluations based on the Pascal VOC~\cite{pascal-voc-2012} standards. However, we impose restrictions on the evaluation sequence or class due to some misalignment issues or overlaps in the annotations related to nighttime labels. We evaluate only the `car' class in the `zurich\_city\_03\_a' sequence. Most methods, except CMDA, achieve satisfactory nighttime translation performance in Bin 1, leading to an overall improvement in performance. However, for bin 3, except for our method, all others could not avoid setbacks. In particular, EventGAN, an image-based generation method, exhibits the most significant decline across multiple bins as it fails to generate accurate information. In contrast, our method demonstrates higher accuracy than the day event scenario across all bins. With the enlargement of the bin size, temporal rules become essential, unlike other methods grappling with this challenge. The qualitative results for object detection can be observed in~\cref{fig:ob_result}.

\subsection{Generate Clean Day Images using Night Events.} 
\begin{figure}[t]
    \centering\includegraphics[width=0.99\linewidth]{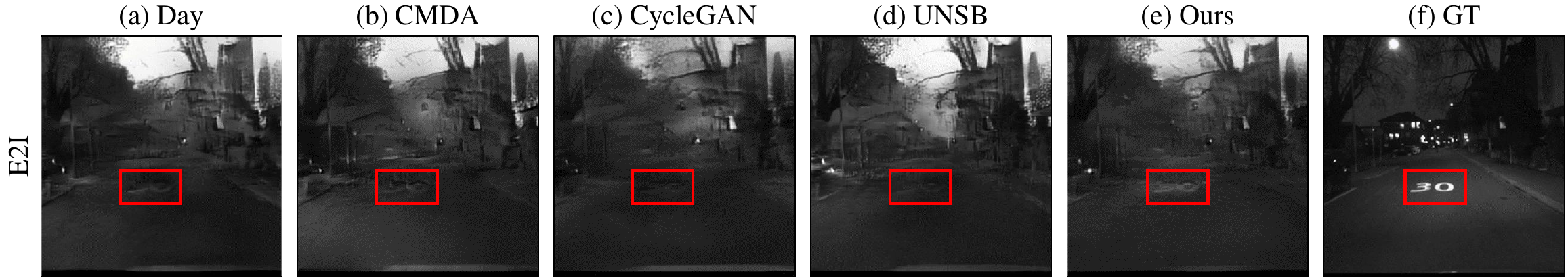}
    \caption{Visual comparison of downstream tasks using real night events.}
    \label{fig:e2i}
\end{figure}

Differences exist in the distributions of events during the day and night. Consequently, a model optimized for day events might encounter difficulties when applied to night events. While the conventional approach involves training the model with annotated data from both day and night, the scarcity of annotated night events poses a challenge. To tackle this limitation, our objective was to transform annotated daytime events into a nighttime format, thereby addressing the shortage of nighttime data. We introduce a specialized network for this purpose. To assess the effectiveness of this strategy in overcoming actual data scarcity, we experiment by using translated night events as training data in a fundamental event model, specifically employing an E2I network. In this process, we substitute certain sequences in the model that normally convert day events to day images with translated night events. Subsequently, we evaluate whether including real night events produced clear images with SSIM in \cref{tab:e2i}. Among them, the model trained with our proposed method performs best in bin 3. We compare the performance of generation clean day images while changing the translation methods in~\cref{fig:e2i}.

\begin{table}[t]
    \centering
    \caption{The translation of night events to an image. SSIM is adopted for the metric.}
    \setlength\tabcolsep{3pt}
    \resizebox{0.99\linewidth}{!}{
        \begin{tabular}{lccccc}
        \hline
        Method & Day & ${\mathrm{v2e}}^{\ast}$~\cite{hu2021v2e} & DVS-${\mathrm{Voltmeter}}^{\ast}$~\cite{lin2022dvs} & ${\mathrm{EventGAN}}^{\ast}$~\cite{zhu2021eventgan} & CMDA~\cite{xia2023cmda} \\ \hline
        SSIM & 0.5180 & 0.5347 & 0.5318 & 0.5314 & 0.5322 \\ \hline
        Method & CycleGAN~\cite{zhu2017unpaired} & GcGAN~\cite{fu2019geometry} & CUT~\cite{park2020contrastive} & UNSB~\cite{kim2023unpaired} & Ours \\ \hline SSIM & 0.5336 & 0.5322 & 0.5315 & 0.5306 & \textbf{0.5428} \\ \hline
    \end{tabular}}
    \label{tab:e2i}
\end{table}

\section{Night-to-Day Event Translation}
\label{sec:night-to-day}
To achieve our research objective, our study aims to translate annotated day events into night events.
We conduct thorough analyses to discern the disparities between day and night events, utilizing our proposed modules to achieve stylistic transfer. 
Meanwhile, we become interested in the converse procedure, translating night events to day events using the same network architecture.
To satisfy this curiosity, we undertake a supplementary experiment by interchanging the input and output of the network.

As a significant contender among comparison groups, we employ  CycleGAN~\cite{zhu2017unpaired} for benchmarking purposes. Our experiments focus on the event histogram of bin 3, and for evaluation, we utilize the FVD score and E2I-FID/KID scores previously employed, with a shift in the control group from real night events to real day events.

\begin{figure}[t!]
    \begin{minipage}[t!h]{.35\linewidth}
        \captionof{table}{Comparison results of night-to-day translations.}
        \renewcommand{\arraystretch}{1.1}
        \resizebox{0.99\linewidth}{!}{
            \centering
            \setlength\tabcolsep{6.2pt}
            \begin{tabular}{lccc}
            \hline
            Method & CycleGAN & Ours  \\ \hline
            FVD$\downarrow$ & 325.47 & \textbf{48.24}\\
            E2I-FID$\downarrow$ & 85.26 & \textbf{78.24} \\
            E2I-KID$\downarrow$ & \textbf{8.54} & 9.68 \\
            \hline
        \end{tabular}}
        \label{tab:day_night}      
    \end{minipage}
    \hfill
    \begin{minipage}[t!h]{.57\linewidth}
        \centering
        \captionof{figure}{Additional experiments of translation from night events to day events.}
        \resizebox{0.99\linewidth}{!}{
        \includegraphics[]{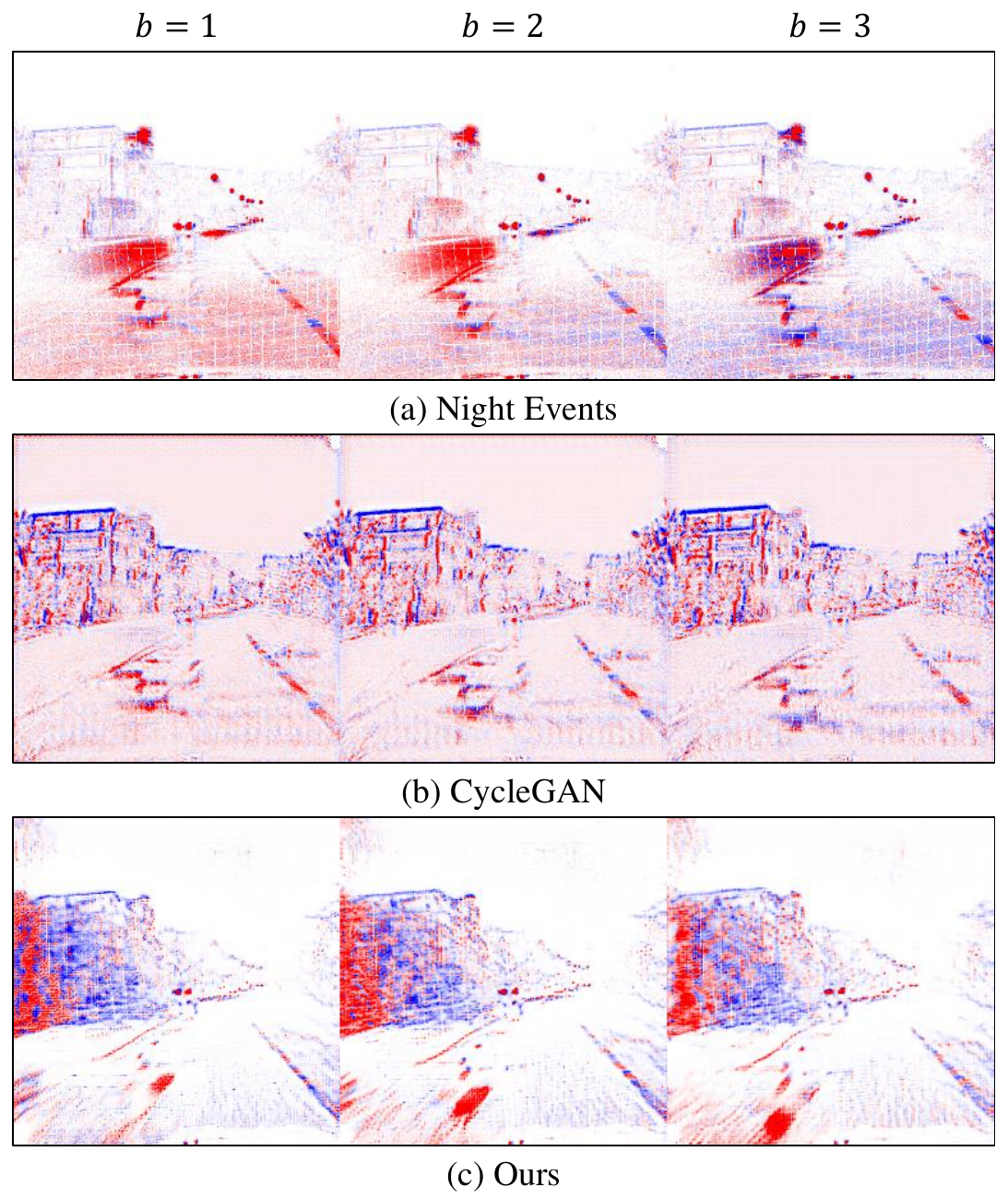}
        \label{fig:day_night}}
    \end{minipage}
\end{figure}

We quantitatively compare our method with CycleGAN in the same manner. In Table~\ref{tab:day_night}, our approach demonstrates significantly superior FVD; for the E2I-FID score, ours show a slightly improved value. Visualizing those as shown in Fig.~\ref{fig:day_night}, our method effectively captures various characteristics of shaping day events, such as noise reduction on the pavement and clearing the background. However, there is room for improvement in terms of shaping the structure. As a result, while there is a significant difference in the FVD score, the E2I-FID/KID scores do not show a substantial distinction. This result demonstrate that our method is able to use in many situations due to abilities of analyzing the events.

\clearpage

%
%

\end{document}